\documentclass{article}

\PassOptionsToPackage{numbers}{natbib}
\usepackage[preprint]{neurips_2022}

\usepackage[utf8]{inputenc} % allow utf-8 input
\usepackage[T1]{fontenc}    % use 8-bit T1 fonts
\usepackage{hyperref}       % hyperlinks
\usepackage{url}            % simple URL typesetting
\usepackage{booktabs}       % professional-quality tables
\usepackage{amsfonts}       % blackboard math symbols
\usepackage{nicefrac}       % compact symbols for 1/2, etc.
\usepackage{microtype}      % microtypography
\usepackage[dvipsnames, svgnames, x11names]{xcolor}
\usepackage{amsthm}
\usepackage{helvet}
\usepackage{courier}
\usepackage{enumerate}
\usepackage{multirow}
\usepackage{array}
\usepackage{graphics}
\usepackage{graphicx}
\usepackage{subfigure}
\usepackage{multirow}
\usepackage{amssymb}
\usepackage{mathrsfs}
\usepackage{extarrows}
\usepackage{booktabs}
\usepackage{setspace}
\usepackage{algorithm}
\usepackage{algorithmic}
\usepackage{bbm}
\usepackage{caption}
\title{A Unified Continuous Learning Framework for Multi-modal Knowledge Discovery and Pre-training}
\author{	
Zhihao Fan, Zhongyu Wei,  Jingjing Chen,  Siyuan Wang, \\ \bf{Zejun Li}, \bf{Jiarong Xu}, \bf{Xuanjing Huang}\\
Fudan University \\
$\{$fanzh18,zywei,chenjingjing,wangsy18,zejunli20,jiarongxu,xjhuang$\}$@fudan.edu.cn}

\begin{document}
\maketitle

\begin{abstract}
    % multi-modal pre-trained model has shows its powerful capability in various downstream tasks including multi-modal knowledge discovery from unstructured unimodal data. In the 
    % Knowledge guided multi-modal pre-training and the knowledge discovery are two important topics to connect the knowledge and multi-modal model. 
    %Pre-trained model and knowledge discovery are two important topics in cross-modal research to bridge perception and cognition. 
Multi-modal pre-training and knowledge discovery are two important research topics in multi-modal machine learning. Nevertheless, none of existing works make attempts to link knowledge discovery with knowledge guided multi-modal pre-training. In this paper, we propose to unify them into a continuous learning framework for mutual improvement. Taking the open-domain uni-modal datasets of images and texts as input, we maintain a knowledge graph as the foundation to support these two tasks. For knowledge discovery, a pre-trained model is used to identify cross-modal links on the graph. For model pre-training, the knowledge graph is used as the external knowledge to guide the model updating. These two steps are iteratively performed in our framework for continuous learning. The experimental results on MS-COCO and Flickr30K with respect to both knowledge discovery and the pre-trained model validate the effectiveness of our framework. 

\end{abstract}

\section{Introduction}
Recent years have witnessed extensive research efforts to align the semantics across vision and language for various tasks, e.g., image-text retrieval~\cite{lin2014microsoft,young2014image,fan2021negative,fan2021constructing}, visual question answering~\cite{antol2015vqa,goyal2017making}, and phrase grounding~\cite{yu2016modeling,plummer2015flickr30k}. In order to break boundaries between tasks, pre-training methods are developed to learn the cross-modal representations based on large-scale training corpora consisting of image-text pairs~\cite{lu2019vilbert,li2019visualbert, su2019vl,li2020oscar,tan2019lxmert,li2020unicoder,zhou2020unified,li2021align,kim2021vilt,zhang2021vinvl,li2022mvp} and the learned model can be adapted to downstream tasks through fine-tuning. 

Owing to the impressive power of the pre-trained model, existing research explores to identify multi-modal knowledge from the unstructured data of images and sentences. VisualSem~\cite{alberts2020visualsem} adopts CLIP~\cite{radford2021learning} to identify semantic concepts from images, and MKGformer~\cite{chen2022hybrid} unifies transformer~\cite{vaswani2017attention} based encoders in different modalities with a hybrid manner for multi-modal knowledge graph completion. In the other direction, researchers also explore to inject external knowledge into the process of multi-modal pre-training. Taking scene graph as external knowledge, ERNIE-ViL~\cite{yu2020ernie} devise the graph masking strategy to utilize intra-modality knowledge in the language side, and ROSITA~\cite{cui2021rosita} extend the strategy to both modalities for fine-grained semantic alignment. These models are competitive and achieve impressive performance in downstream tasks. 

Nevertheless, none of existing works make attempts to link knowledge discovery with knowledge guided multi-modal pre-training. We argue that there are several advantages to link them together. First, it naturally forms a continuous multi-modal learning paradigm which makes full use of training data without any annotation efforts in obtaining the explicitly labeled knowledge graphs (e.g., scene graph). Second, the knowledge graph could help to guide the pre-training process in return, which enables better performances for downstream tasks. 

\begin{figure}[!th]
\centering
\includegraphics[width=1.0\textwidth]{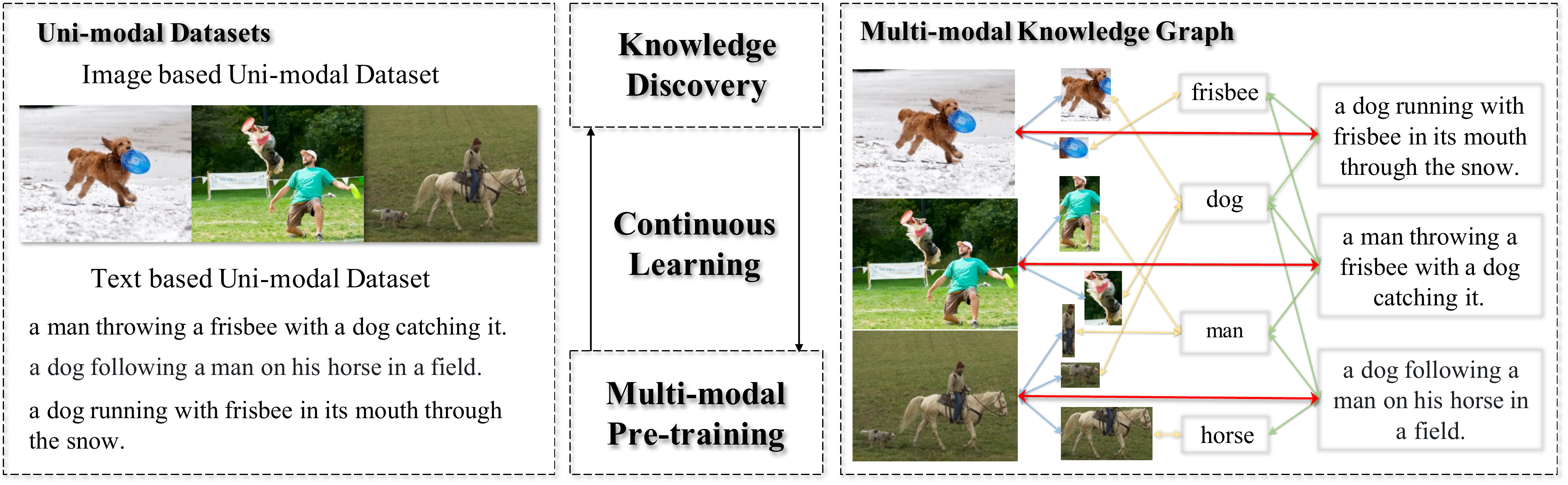}
\centering
\caption{The unified framework which unifies knowledge discovery and multi-modal pre-training.}
\label{intro_example}
\end{figure}

To this end, we propose to integrate knowledge discovery and knowledge guided multi-modal pre-training into a unified continuous learning framework in this paper. The overall architecture is presented in Figure~\ref{intro_example}. We take open-domain unstructured data of images and texts as input to align the two modalities. A multi-modal knowledge graph is maintained as the foundation of our framework to support knowledge discovery and model pre-training. For knowledge discovery, a pre-trained model is used to extract knowledge units from the uni-modal datasets and update the knowledge graph by linking multi-modal semantic units. For model pre-training, the knowledge graph is used as the external knowledge to guide the updating of the multi-modal model. %with sampled knowledge units. 

In practice, the multi-modal graph includes three kinds of nodes to cover multiple levels of semantic units, namely, image, sentence and concept (objects in the vision side and phrases in the language side). Knowledge discovery is then treated as a link prediction task on the knowledge graph taking pre-training model as a probe. Model training is then guided by the links predicted in the knowledge discovery process. These two steps are performed iteratively for continuous learning. In the optimization, we design confidence score aware soft labels and add regularization to mitigate the problem of error propagation.

For knowledge discovery, we evaluate the quality of edges identified in the multi-modal knowledge graph. For the pre-trained model, we evaluate its performance on one typical downstream task, i.e., image-text retrieval. Experiments on MS-COCO~\cite{lin2014microsoft} and Flickr30K~\cite{plummer2015flickr30k} demonstrate the effectiveness of our proposed framework. We also configure the noisy uni-modal datasets to validate the robustness of our model and results show that the framework can perform reliable model training even in the noisy environment.

\section{Related Work}
% \paragraph{Vision-Language Representation Learning}
\textbf{Vision-Language Representation Learning.}\ \ % Through training on the human-annotated image-sentence pairs, the knowledge of these inter-modality linking is injected to the multi-modal model. 
Following BERT~\cite{devlin2018bert}, UNITER~\cite{chen2020uniter}, Unicoder~\cite{li2020unicoder}, ERNIE-ViL~\cite{yu2020ernie} and VinVL~\cite{zhang2021vinvl} pre-train the vision-language transformer model on the large-scale annotated image-sentence pairs to learn the knowledge of the inter-modality links. In language pre-training, explicitly incorporating external knowledge to enhance model pre-training is investigated by ERNIE~\cite{sun2019ernie,zhang2019ernie} and widely explored in recent years~\cite{liu2020k,wang2020k,wang2021kepler}. The knowledge facilitates model pre-training by an improved structural masked language model to enhance the modeling in structure, such as syntactic and semantic. In the vision-language pre-training, ERNIE-ViL~\cite{yu2020ernie} and ROSITA~\cite{cui2021rosita} integrate the knowledge of concepts from image and sentence to enhance the semantic alignments. In our work, we introduce a multi-modal knowledge graph in the dataset where the concepts, namely, phrases and objects, are unified.

\textbf{Knowledge Discovery via Link Prediction.}\ \ The construction of knowledge graph aims to predict the missing relation or entity in the triple~\cite{onoro2017answering,liu2019mmkg,pezeshkpour2018embedding,shah2019open,hamann2021open}. In multi-modal knowledge graph,~\cite{liu2019mmkg} constructs datasets to predict the multi-relational links between entities associated with numerical and visual data. VisualSem~\cite{alberts2020visualsem} adopts CLIP~\cite{radford2021learning} to identify semantic concepts from images. To test the model's generalization ability,~\cite{hamann2021open} presents an open-world knowledge graph completion model to predict the links among entities unseen in training set. These works concentrate on knowledge discovery with the fixed multi-modal model. Different from these works, we perform knowledge discover and multi-modal model training in a iterative way, thus the knowledge graph is obtained with the progressively improved multi-modal model. 
\textbf{Pseudo Labeling}\ \ Given the class labels, pseudo-labeling~\cite{lee2013pseudo,shi2018transductive} aims to pick a pseudo one for unlabeled samples with a model trained on labeled data.~\cite{lee2013pseudo} produces pseudo-labels from the prediction of a trained model.~\cite{iscen2019label} assigns labels for unlabeled samples through the propagation on the graphs.~\cite{shi2018transductive} incorporates confidence scores for unlabeled samples on the density of a local neighborhood.~\cite{rizve2021defense} considers the uncertainty of the confidence score for better network calibration. In the classification based pseudo labeling, the label is determined by the threshold. The same method in link assignment leads to popular nodes dominate the graph which degrades the graph quality and hinders the training of nodes with less edges. Thus, we estimate the popularity and devise the label-aware threshold for link assignment. 

%label-aware link assignment strategy.

\section{Continuous Learning Framework}

The workflow of our continuous learning framework is presented in Algorithm.~\ref{algorithm-1}. Given two uni-modal datasets (images and texts) and an initialized version of multi-modal pre-training model, we aim to deliver a multi-modal knowledge graph and an updated version of the pre-training model. In each iteration $t$, the knowledge graph and the pre-training model can be updated through knowledge discovery and model training. 

%On the basis of $F_{\theta}$, we perform link prediction between image $I$ and sentence $T$, then object $O\in I$ and phrase $P\in T$. In timestamp $t$, through the pseudo link $\hat{L}^{t}_{(I,T)}$ in global level and $\hat{L}^{t}_{(O,P)}$ in local level, we build a multi-modal graph $\mathcal{G}^{t}$ to present the discovered knowledge. The continuous learning process is shown in Algorithm.~\ref{algorithm-1}. Each iteration in timestamp $t$ has two steps: (1) Knowledge discovery with multi-modality graph: Assignment of pseudo links with the updated model $F_{\theta}^{t-1}$ to construct the new graph $\mathcal{G}^{t}$. (2) multi-modal model training: Training the model with the knowledge in the graph $\mathcal{G}^{t}$.

\subsection{Task Preliminary}

\textbf{Uni-modal Datasets.}\ \ Given two Uni-modal datasets of images $\mathcal{I}$ and texts $\mathcal{T}$, we construct a knowledge graph and train the multi-modal pre-training model.

\textbf{Multi-modal Knowledge Graph.} \ \ We maintain a graph $\mathcal{G}$ to store the multi-modal knowledge. It has four kinds of nodes: image $I\in\mathcal{I}$ and sentence $T\in\mathcal{T}$ in global level, object $O$ and phrase $P$ in local level. It has two kinds of edges: intra-modality ones $(L_{(O,I)}$, $L_{(P,T)})$ and inter-modality ones $(L_{(I,T)}$, $L_{(O,P)})$. For intra-modality edges, we have link $L_{(O,I)}=1$ if $O\in I$ and $L_{(P,T)}=1$ if $P\in T$. 

\textbf{Knowledge Discovery.}\ \ We identify cross-modality edges, i.e., $\big\{L_{(I,T)},L_{(O,P)}\big\}$, in the multi-modal knowledge graph for knowledge discovery. This starts with a multi-modality pre-training model $F_{\theta}$, and it assigns confidence scores in $[0, 1]$ to cross-modality edges as Eq.~(\ref{target_function}).
%In our setting, the ground-truth inter-modality links $\big\{L_{(I,T)},L_{(O,P)}\big\}$ are unknown. To estimate the linking possibility $\mathbb{P}\big(L_{(I,T)}=1\big|I,T\big)$ and $\mathbb{P}\big(L_{(O,P)}=1\big|O,P\big)$, a model $F_{\theta}$ which is initialized from pre-training, e.g., VinVL~\cite{zhang2021vinvl}, is employed to output the linking confidence score in $[0, 1]$ as Eq.~(\ref{target_function}).
\begin{equation}
    \begin{normalsize}
        \begin{gathered}
            F_{\theta}(I,T)\rightarrow[0,1],\ \ F_{\theta}(O, P)\rightarrow[0,1],\ \textrm{where}\ O\in I,\ P \in T \label{target_function}
        \end{gathered}
    \end{normalsize}
\end{equation}
 
\textbf{Multi-modal Pre-training.}\ \ We sample cross-modality edges ($\{\hat{L}^{t}_{(I,T)}, \hat{L}^{t}_{(O,P)}\}$) from the multi-modal knowledge graph as supervisions for continuous multi-modal pre-training.

\begin{algorithm}[!t]
% \setstretch{1.0}
\caption{The overall procedure of our framework.}
\label{alg:algorithm}
% \hspace*{\algorithmicindent} 
\textbf{Input}: Unimodal datasets $\mathcal{I}$ and $\mathcal{T}$. A pre-trained model $F_{\theta}$. The maximal iteration step $t_{max}$. 
% \hspace*{\algorithmicindent} 
\textbf{Output}: The multi-modality graph $\mathcal{G}^{t_{max}}$ and the multi-modal model $F_{\theta}^{t_{max}}$.

\begin{algorithmic}[1]
\STATE Initialize $F_{\theta}^{0}$ with $F_{\theta}$ and $t=1$.
\WHILE{$t\leq t_{max}$}
\STATE {\color{DarkGreen}{// Knowledge discovery as link prediction on multi-modal graph.}}
% \FOR{$I\in\mathcal{I}$, $T\in\mathcal{T}$}
% \STATE Compute the popularity $E(I)$ and $E(T)$ following Eq.~(\ref{estimate}).
% \STATE Compute the class-aware threshold $\lambda(I)$ and $\lambda(T)$. 
% \ENDFOR
\FOR{$(I,T)\in\mathcal{I}\times\mathcal{T}$}
\STATE {\color{DarkGreen}{// Global level link prediction.}}
\STATE Compute the class-aware threshold $\lambda(I)$ and $\lambda(T)$ following Eq.~(\ref{estimate}) and Eq.~(\ref{improved_version_pseudo_linking}). 
\STATE $\hat{L}^{t}_{(I,T)}=1/2\times\big(\mathbbm{1}\big[F_{\theta}^{t-1}(I,T)\ge\lambda(I)\big]+\mathbbm{1}\big[F_{\theta}^{t-1}(I,T)\ge\lambda(T)\big]\big)$
%\IF{$\hat{L}^{t}_{(I,T)}\ne 0$}
\FOR{$(O,P)\in I\times T$} 
\STATE {\color{DarkGreen}{// Local level link prediction.}}
\STATE $ \hat{L}^{t}_{(O,P)}=\mathbbm{1}\big[F^{t-1}_{\theta}(I,T)\ne 0\big]\mathbbm{1}\big[F^{t-1}_{\theta}(O,P)\ge \lambda_{C}\big] $ % Assignment of pseudo link $\hat{L}^{t}_{(O,P)}$ following Eq.~(\ref{concept_label_assignment}).
\ENDFOR
% \ENDIF
\ENDFOR
% \STATE Construction of $\mathcal{G}^{t}$ finishes.
% \FOR{$(I,T)\in\mathcal{I}\times\mathcal{T}$ and $\hat{L}^{t}_{(I,T)}\ne 0$}
% \FOR{$(O,P)\in I\times T$}
% \STATE $ \hat{L}^{t}_{(O,P)}=\mathbbm{1}\big[F^{t-1}_{\theta}(O,P)\ge \lambda_{C}\big] $ % Assignment of pseudo link $\hat{L}^{t}_{(O,P)}$ following Eq.~(\ref{concept_label_assignment}).
% \ENDFOR
% \ENDFOR
\STATE {\color{DarkGreen}{// multi-modal pre-training.}}
\FOR{$I\in\mathcal{I}$}
\STATE {\color{DarkGreen}{// Graph structure enhanced representation learning.}}
\STATE {\color{DarkGreen}{// Collect phrase nodes of image neighbors.}}
\STATE $\mathcal{P}^{t}_{I}=\big\{P|P\in T, O\in I, \hat{L}^{t}_{(I,T)}=1, \hat{L}^{t}_{(O,P)}=1\big\}$
\STATE {\color{DarkGreen}{// Confidence score aware training.}}
% \STATE {\color{DarkGreen}{// Image-sentence level loss.}}
\STATE Sample $T^{+}$ from $\{T\big|\hat{L}^{t}_{(I,T)}=1\}$ if it is not empty, otherwise from $\{T|\hat{L}^{t}_{(I,T)}=0.5\}$
\STATE Sample $T^{-}$ from $\{T\big|\hat{L}^{t}_{(I,T)}=0\}$
\STATE Construct soft labels $Y_{(I,T^{+})}^{t}$ and $Y_{(I,T^{-})}^{t}$ for $T^{+}$ and $T^{-}$, respectively. 
\FOR{$T\in\{T^{+},T^{-}\}$}
\STATE Model forward with $F_{\theta}(I,T,\mathcal{P}^{t}_{I})$.
\STATE Compute global level loss in Eq.~(\ref{image-sentence-linking-prediction}), local level loss in Eq.~(\ref{concept_level_loss}), and regularization loss in Eq.~(\ref{uncertainty_loss}).
\ENDFOR
\ENDFOR
\STATE Repeat the similar training procedure (line $14\sim26$) for each $T\in\mathcal{T}$.
\STATE $F^{t}_{\theta}:=F_{\theta}$, $t:=t+1$.
\ENDWHILE
% \ENDFOR
\end{algorithmic}
\label{algorithm-1}
\end{algorithm}

%\section{Continuous Learning Framework}

% Each iteration loop of the system has two steps: (1) knowledge discovery with multi-modality graph: Filtering noisy pseudo links with the updated model to adjust the old graph. (2) multi-modal model training: Training the model with the graph based knowledge. 

% To build the cross-modal graph on the basis of the unimodal datasets, we propose to progressively adding pseudo linking as time goes on.
% For continuous knowledge discovery from the unimodal datasets, we propose to train the pre-trained model with the pseudo multi-modal graph. Through this, we not only store the graph knowledge, but also can progressively extend the knowledge graph with the improved generalization capability of the pre-trained model. Following the configuration of pseudo linking, the task of link prediction is also composed of two stages, namely, image-sentence level and concept level. Before the description of the detailed link prediction tasks, we first introduce the pre-trained model enhanced with pseudo graph.

\subsection{Knowledge Discovery as Link Prediction on Multi-modal Graph}
% \subsection{Progressive Graph Construction}
We treat knowledge discovery as the task of link prediction on the multi-modal graph on two semantic levels, namely, global level (image and sentence) and local level (object and phrase). 

% The key challenge of the task is that none of labels is available, thus pseudo linking needs to be assigned in the cross-modal graph. 
% In the multi-modal graph, we expect nodes of image and sentence gather as cluster centered on an image or sentence, the semantic discrepancy of different clusters makes nodes not in the same cluster disconnected.
\subsubsection{Global Level Link Prediction} 
\label{Global_Level_Link_Prediction}
Given a node pair that contains an image $I$ and a sentence $T$, the pre-trained model is employed to compute a confidence score to determine the existence of the edge. This is regarded as a classification problem and a threshold $\lambda$ is introduced to make the decision~\cite{lee2013pseudo,shi2018transductive}. The calculation is shown in Eq.~(\ref{base_version_pseudo_linking}).
\begin{equation}
    \begin{normalsize}
        \begin{gathered}
            \hat{L}^{t}_{(I,T)}=\mathbbm{1}\big[F_{\theta}^{t-1}(I,T)\ge \lambda\big] \label{base_version_pseudo_linking}
        \end{gathered}
    \end{normalsize}
\end{equation}
%based on their representations learned by the pre-training model. 
%as the anchor, the pseudo link $\hat{L}^{t}_{(I, T)}$ to sentence $T$ is determined through the confidence score, so is for sentence as anchor. Following the setting of most vision-language pre-trained models, we compute the confidence score of the image sentence pair with the \textrm{[CLS]} output representation, and regard the issue as a binary classification problem. 
%In previous classification-based pseudo label assignment~\cite{lee2013pseudo,shi2018transductive}, an absolute threshold $\lambda$ is employed to compare with the confidence score of each class to determine the pseudo label as Eq.~(\ref{base_version_pseudo_linking}), which aims to reserve the accurate labels and filter the noisy ones. % The basic version of pseudo link assignment can be described in the following 
% discard

\paragraph{Label-aware Link Assignment (LA).} After link prediction on the graph, images are labeled with connected sentences, or vice versa. Following this interpretation, we can regard $I$ as input and $T$ as label in Eq.~(\ref{base_version_pseudo_linking}). Some nodes (image or sentences) in the graph might be more popular than others. Nodes with less edges suffer from the problem of under-training. And links connected to these kinds of nodes tend to obtain lower confidence scores. To mitigate this problem, we propose to set the threshold dynamically for a link considering the popularity of nodes (also known as labels) it connects to. First, we estimate the confidence score $\mathbb{P}\big(L=1\big|I\big)$ of the label to measure its popularity $E(I)$ as Eq.~(\ref{estimate}). We assume sentence $T$ is sampled from an empirical (uniform) distribution and $\mathbb{P}\big(T|I\big)$ is thus constant. %In implementation, we sample these sentences which earn top-tiers for computation.
% Let Cx denote the potential imputed label for image x, i.e., Cx = arg maxy P(y|x; θ). We use a class-aware threshold $\lambda_{I}$ for image as:
\begin{equation}
    \begin{normalsize}
    \mathbb{P}\big(L=1\big|I\big)=\sum_{T}\mathbb{P}\big(L=1\big|T,I\big)\mathbb{P}\big(T|I\big)\approx\frac{1}{\big|\{T_{k}\}\big|}\sum_{T_{k}}\mathbb{P}\big(L=1\big|T_{k},I\big)=E(I) \label{estimate}
    \end{normalsize}
\end{equation}
In addition, we add the power hyperparameter $\mu$ to compute the corresponding label-aware threshold $\lambda(I)$. So is for sentence $T$.
\begin{equation}
    \begin{normalsize}
        \begin{gathered}
            \lambda(I)=\big[E(I)\big]^{\mu} \label{improved_version_pseudo_linking}
        \end{gathered}
    \end{normalsize}
\end{equation}

\paragraph{Bi-label Link Assignment (BL).} For a cross-modality edge, both nodes it connects obtain popularity scores. Thus, we propose to consider thresholds of both nodes for link prediction and classify links into two categories, namely strong and weak ones. If the confidence score of a link is higher than both thresholds, it is treated as a strong one. If the score is only higher than one threshold, it is a weak one. We denote $\hat{L}^{t}_{(I, T)}=0.5$ for weak links. Overall, the links between image $I$ and sentence $T$ can be summarized as Eq.~(\ref{pseudo_links}).
\begin{equation}
    \begin{normalsize}
        \begin{gathered}
            \hat{L}^{t}_{(I,T)}=\frac{1}{2}\times\big(\mathbbm{1}\big[F_{\theta}^{t-1}(I,T)\ge\lambda(I)\big]+\mathbbm{1}\big[F_{\theta}^{t-1}(I,T)\ge\lambda(T)\big]\big) \label{pseudo_links}
            % \hat{L}^{t}_{(I,T)}= \left\{\begin{array}{ll}
            %      1, &\textrm{if}\ \mathbbm{1}\big[F_{\theta}^{t-1}(I,T)\ge \mathop{\textrm{max}}\big(\lambda(I), \lambda(T)\big)\big]=1 \vspace{1ex}\\ 
            %      0.5, &\textrm{if}\ \mathbbm{1}\big[F_{\theta}^{t-1}(I,T)\ge \mathop{\textrm{min}} \big(\lambda(I), \lambda(T)\big)\big]=1\vspace{1ex}\\ 
            %      0, & \textrm{otherwise}\label{pseudo_links}
            % \end{array}\right.
        \end{gathered}
    \end{normalsize}
\end{equation}

% \begin{equation}
%     \begin{normalsize}
%         \begin{gathered}
%             \hat{L}^{t}_{(I,T)}= \left\{\begin{array}{ll}
%                  1, &\textrm{if}\ \mathbbm{1}\big[F_{\theta}^{t-1}(I,T)\ge \lambda\big]\mathbbm{1}\big[R_{I}(T)\le \lambda_{I}\big]\mathbbm{1}\big[R_{T}(I)\le \lambda_{T}\big]=1 \vspace{1ex}\\ 
%                  l, &\textrm{if}\ \mathbbm{1}\big[F_{\theta}^{t-1}(I,T)\ge \lambda\big]\big(\mathbbm{1}\big[R_{I}(T)\le \lambda_{I}\big]+\mathbbm{1}\big[R_{T}(I)\le \lambda_{T}\big]\big)=1\vspace{1ex}\\ 
%                  0, & \textrm{otherwise}\label{pseudo_links}
%             \end{array}\right.
%         \end{gathered}
%     \end{normalsize}
% \end{equation}

% \begin{equation}
%     \begin{normalsize}
%         \begin{gathered}
%             T\in\mathop{\textrm{topk}}\big(I, \big\{T_{k}\big\}, n^{I}\big),\ I\in \mathop{\textrm{topk}}\big(T, \big\{I_{k}\big\}, n^{T}\big)
%         \end{gathered}
%     \end{normalsize}
% \end{equation}

\subsubsection{Local Level Link Prediction} 
\label{section_concept_level_linkig_assignment} 
For the local level, we link concepts in different modalities. For computation efficiency, we only consider links between objects and phrases whose global nodes (image and sentence) are connected in the graph. On the language side, we utilize spacy~\footnote{https://spacy.io/} for noun phrases identification and filter out those with low frequency to form a phrase set. On the vision side, an object detector is used to locate objects~\cite{girshick2015fast}. Following~\cite{dou2021improving}, the representation of phrase is computed through mean-pooling of its tokens, then we measure the cosine similarity between the phrase and all objects in image $I$. Softmax is utilized to normalize the similarity scores. At last, the object whose similarity score is larger than $\lambda_{C}$ would be linked to the phrase. 
\begin{equation}
    \begin{normalsize}
        \begin{gathered}
            % \hat{L}^{t}_{(O,P)}=\hat{L}^{t}_{(I,T)}\mathbbm{1}\big[F^{t-1}_{\theta}(O,P)\ge \lambda_{C}\big]
            \hat{L}^{t}_{(O,P)}=\mathbbm{1}\big[F^{t-1}_{\theta}(I,T)\ne 0\big]\mathbbm{1}\big[F^{t-1}_{\theta}(O,P)\ge \lambda_{C}\big] \label{concept_label_assignment}
        \end{gathered}
    \end{normalsize}
\end{equation}

%$ \subsection{Continuous Model Training}
\subsection{Multi-modal Pre-training}
Based on the constructed multi-modal knowledge graph, we sample cross-modality links for model training. Links in the graph are identified automatically without human monitoring, therefore it is likely to contain noisy supervision. In order to mitigate the problem of error propagation, we propose a confidence score aware mechanism for model pre-training. We use image-text matching as the pre-training task. Note that, our framework is compatible to include other pre-training tasks. 

\subsubsection{Graph Structure Enhanced Representation Learning}
% Through the construction of graph $\mathcal{G}^{t}$ in time-stamp $t$, we find out neighbors from the other modality for each node in its own cluster, especially the image node. These sentence nodes as the neighbors of the image node are in the textual corpus can help bridge the cross-modal gap when perform inter-modality link prediction. Thus, in the following modeling, we take the image node as central node, and aggregate the sentence neighbors for representation learning. 

For an input image or sentence, we learn their representations taking the graph structure into consideration. For each node (image or sentence), two-hop neighbors are considered. Take image node as an example, the directly linked sentences and phrase nodes connected to these sentences are included. Therefore, we integrate multiple levels of semantic information into the learning process. In practice, the linked sentences $\mathcal{N}_{I}=\{T|\hat{L}^{t}_{(I,T)}=1\}$ and corresponding phrases $\mathcal{P}^{t}_{I}=\{P|P\in T, O\in I, \hat{L}^{t}_{(I,T)}=1, \hat{L}^{t}_{(O,P)}=1\}$ are concatenated to the original image to form an input sequence. The phrases share the token embedding with the sentence, and we add a new segment embedding vector to denote that it shares the semantics with the image instead of the sentence. % as Eq.~(\ref{new_graph_modeling}).  
% \begin{equation}
%     \begin{normalsize}
%         \begin{gathered}
%             F_{\theta}(I,T)\stackrel{\mathcal{G}^{t}}{\Longrightarrow} F_{\theta}(I, \mathcal{P}^{t}_{I},T) \label{new_graph_modeling}
%         \end{gathered}
%     \end{normalsize}
% \end{equation}
% we propose to utilize the concept-level linking to fully exploit the constructed pseudo graph. 

% The multi-modal graph sets up the association across different modalities. We propose to utilize these linking to better bridge semantic gap between different modalities. 

% During model training, motivated by the research about the role of local neighbors in link prediction [], we combine the linked nodes of phrases into image to perform graph encoding. Phrases are in the same modality with sentence and share the semantics of the image, thus they can contribute to better link prediction.

\subsubsection{Confidence Score Aware Training}
Instead of directly using hard labels as supervision we use confidence scores of links to construct soft labels for training. Losses are computed in both global (image-text) and local (phrase-object) levels. 

%It is inevitable that the error would exist in $\mathcal{G}^{t}$. In this part, we devise the noise-resistant loss for link prediction on the basis of the linking confidence score.

\label{section_linking_prediction_loss}
\paragraph{Global Level Loss.} 

%Link prediction in image-sentence level is similar to the image-text retrieval~\cite{NIPS2013_7cce53cf}. We need to assign positive and negative sentences (images) for the query image (sentence). 

There are three types of cross-modality edges between nodes according to the confidence score, namely, strong one, weak one and none. We construct a triplet for loss computation. Take the image as an example. For each image $I$, we randomly select a linked sentence node $T^{+}\in\{T| L^{t}_{(I,T)}=1\}$ to compose a positive pair. If there is no $T\in\mathcal{T}$ satisfies the requirement, the sentence is sampled from the weak linked sentences $T^{+}\in\{T|L^{t}_{(I,T)}=0.5\}$. The negative sentence $T^{-}$ is sampled from $\{T|L^{t}_{(I,T)}=0\}$ that do not link to image $I$. 

%If we assign hard label $1$ for $T^{+}$ in loss computation, when the link between $T^{+}$ and $I$ is mistakenly assigned, as a result, the optimized confidence score of $T^{+}$ is enlarged. The model thus overfits to noise. The phenomenon occurs because when the confidence score of the linking is low, the model still regards it as a certain positive objective for optimization. This disparity causes the model hard to accommodate these links with low confidence scores. To address this issue, instead of adopting hard label, we utilize the confidence score to construct a soft target label for training. 

For these sampled links, we sharpen the confidence score by the exponential operation with hyperparameter $\gamma$. We observe that weak links are more likely to be mistakenly assigned than strong ones, another hyperparameter $\mu<1$ is thus adopted on the group of weak ones. The label construction is shown in Eq.~(\ref{pseudo_link_labels}).
\begin{equation}
    \begin{normalsize}
        \begin{gathered}
            Y^{t}_{(I,T)}= \left\{\begin{array}{ll}
                 \big[F^{t-1}_{\theta}(I,T)\big]^{\gamma}, &\hat{L}^{t}_{(I,T)}=1 \vspace{0.8ex}\\ 
                 \mu\big[F^{t-1}_{\theta}(I,T)\big]^{\gamma}, &\hat{L}^{t}_{(I,T)}=0.5\vspace{0.8ex}\\ 
                 1-\big[1-F^{t-1}_{\theta}(I,T)\big]^{\gamma}, &\hat{L}^{t}_{(I,T)}=0 \label{pseudo_link_labels}
            \end{array}\right.
        \end{gathered}
    \end{normalsize}
\end{equation}

% Considering that the positive image-sentence pair is likely to be false positive, we propose to keep the uncertainty for them instead of hard label. Combining this with binary cross-entropy, we assign hard label $Y_{(I,T^{-})}=0$ for $(I,T^{-})$, and $Y_{(I,T^{+})}=1$ for $(I,T^{+})$ if $\hat{L}^{t}_{(I,T^{+})}=1$ otherwise $Y_{(I,T^{+})}=y_p\in(0,1)$ where $y_{p}$ is the hyperparameter. 

The loss is shown in Eq.~(\ref{image-sentence-linking-prediction}).
%, $P_{(I,T)}$ outputs the confidence score of the linking $(I,T)$.
\begin{equation}
    \begin{normalsize}
        \begin{gathered}
            % L_{IT}=-\sum_{T\in \{T^{+}, T^{-}\}}\Big[Y^{t}_{(I,T)}\mathop{\textrm{log}}F_{\theta}(I,T)+\big[1-Y^{t}_{(I,T)}\big]\mathop{\textrm{log}}\big[1-F_{\theta}(I,T)\big]\Big] \label{image-sentence-linking-prediction}
            \mathcal{L}_{IT}=-\sum_{(I,T)}Y^{t}_{(I,T)}\mathop{\textrm{log}}F_{\theta}(I,T) \label{image-sentence-linking-prediction}
        \end{gathered}
    \end{normalsize}
\end{equation}

% With the triple $(I, T^{+}, T^{-})$, we can compute the linking possibility $p(I,T^{+})$ and $p(I,T^{-})$ by means of the pre-trained model. To utilize binary cross-entropy for optimization, we assign label $y(I,T^{-})=0$ for $(I,T^{-})$.  and $y(I,T^{+})=1$ for $(I,T^{+})$

\paragraph{Local Level Loss.}
% The concept level link prediction is similar to the task of phrase grounding. 
For the linked pairs of objects and phrases, we regard the phrase $P$ as the anchor and collect all of $O\in I$ for the cross-entropy computation.  In this level, we simply take $Y^{t}_{(O,P)}=\hat{L}^{t}_{(O,P)}$ and optimize with the following Eq.~(\ref{concept_level_loss}).
% to avoid the model linking the phrase to some mismatched object over-confidently, we unify these objects that linked to the phrase $P$ as an entirety through summing up the corresponding normalized scores, and we also add the soft label of pseudo image-sentence linking to the concept level linking in case that the image and sentence that they originate from are linked mistakenly.
% For the pair of image and sentence $(I,T)$ which are linked in $G$, 
\begin{equation}
    \begin{normalsize}
        \begin{gathered}
            \mathcal{L}_{C}=-\sum_{(I,T)}Y^{t}_{(I,T)}\sum_{(O,P)\atop O\in I,P\in T}Y^{t}_{(O,P)}\mathop{\textrm{log}}F_{\theta}(O,P) \label{concept_level_loss}
        \end{gathered}
    \end{normalsize}
\end{equation}

\paragraph{Regularization to Reduce Uncertainty of Confidence.}
% The reliable training on the noisy graph depends on the
%Previous works~\cite{rizve2021defense} empirically shows that the high uncertainty of the confidence score misleads the model to produce incorrectly pseudo labels. 

To ensure the reliability of the linking confidence scores, we further reduce the uncertainty during the training. MC Dropout~\cite{gal2016dropout} is employed for uncertainty estimation. For each input pair $(I,T)$, we perform forward computation twice of the model with different dropout and get the confidence scores $F_{\theta,1}(I, T)$ and $F_{\theta,2}(I, T)$ which can be regard as the two bernoulli distributions,  then utilize the Jensen–Shannon divergence to minimize the distance which is shown in Eq.~(\ref{uncertainty_loss}). 
\begin{equation}
    \begin{normalsize}
        \begin{gathered}
            \mathcal{L}_{U}=\frac{1}{2}\mathop{\textrm{KL}}\big(F_{\theta,1}(I, T)\big\|F_{\theta,2}(I, T)\big)+\frac{1}{2}\mathop{\textrm{KL}}\big(F_{\theta,2}(I, T)\big\|F_{\theta,1}(I, T)\big)\label{uncertainty_loss}
        \end{gathered}
    \end{normalsize}
\end{equation}

% The confidence score is the basis of our model training in the noisy environment.

% To ensure the r

% Previous works~\cite{rizve2021defense} empirically shows that the high uncertainty of the confidence score misleads the model to produce incorrectly pseudo labels, this motivates us to reduce the uncertainty to increase the reliability of the link prediction. In implementation, 

The overall training loss of our model has three components as Eq.~(\ref{overall_loss}) with hyperparameters $\lambda_{\emph{IT}}$, $\lambda_{\emph{C}}$, and $\lambda_{\emph{U}}$.
\begin{equation}
    \begin{normalsize}
    % \begin{small}
        \begin{gathered}
            \mathcal{L}=\lambda_{\emph{IT}}\mathcal{L}_{\emph{IT}}+\lambda_{\emph{C}}\mathcal{L}_{\emph{C}}
            +\lambda_{\emph{U}}\mathcal{L}_{\emph{U}}
            \label{overall_loss}
        \end{gathered}
    % \end{small}
    \end{normalsize}
\end{equation} 

\section{Experiment}
\label{experiment_section}

\subsection{Experiment Setup}
\label{section_experiment_setup}
%To validate the effectiveness of the proposed framework, we evaluate on the acquired knowledge graph and two vision-language tasks, namely, image-text retrieval and phrase grounding under transductive setting. 
% For the acquired knowledge, we evaluate in image-sentence level. For the multi-modal model, we evaluate on two classic vision-language tasks, namely, image-text retrieval and phrase grounding. 
% To validate the effectiveness of the proposed framework, we evaluate both of knowledge discovery and the multi-modal model. For the acquired knowledge, we evaluate in image-sentence level. For the multi-modal model, we evaluate on two classic vision-language tasks, namely, image-text retrieval and phrase grounding. 
To validate the effectiveness of the proposed framework, we evaluate the performance on two tasks, knowledge discovery and cross-modal image-text retrieval under transductive setting. For knowledge discovery, we evaluate the performance of link prediction for both global level (image and sentence) and local level (object and phrase). 

\textbf{Dataset.}\ \ The evaluation is performed on two benchmark datasets, MS-COCO~\cite{lin2014microsoft} and Flickr30K~\cite{young2014image}. 
In test split, MS-COCO and Flickr30K contain 5,000 and 1,000 images respectively, and each image is annotated with 5 text descriptions. In addition to text descriptions, Fickr30K also provides bounding box annotation for each phrase in the description, which enables local level link prediction. In total, there are 16,576 pairs of phrase and object. 

\textbf{Evaluation Metrics.} For knowledge discovery, we follow~\cite{yang2015evaluating} and use recall (R), precision (P) and F1 (F) as evaluation metrics for global level link prediction; while for local level link prediction, since it is equivalent to phrase grounding task, we follow~\cite{wang2021improving} and use the accuracy as the evaluation metric. The accuracy is defined as the fraction of query phrases whose predicted bounding box overlaps ground-truth box with IoU larger than 0.5. For image-text retrieval, We report recall at K (R@K) following the metrics in~\cite{NIPS2013_7cce53cf},
% The accuracy is calculated on the candidate bounding boxes whose IoU with the ground-truth regions are larger than 0.5.
%The multi-modal model is evaluated with image-text retrieval and phrase grounding. We report recall at K (R@K) following the metrics in image-text retrieval~\cite{NIPS2013_7cce53cf}, and the accuracy of the hit in phrase grounding~\cite{yu2016modeling,plummer2015flickr30k} where a hit means the IoU of the ground-truth region and the candidate is greater than 0.5.

\textbf{Implementation.}\ \ We employ VinVL(base)~\cite{zhang2021vinvl} as our backbone. In our continuous learning framework, we need to initialize the multi-modal graph for an iterative optimization. As VinVL~\cite{zhang2021vinvl} (base) are not pre-trained with image-text matching task, CLIP is used to produce the initial linking confidence score between image and text for graph initialization. The hyperparameters are selected according to the results on randomly sampled 100 image-sentence pairs in validation set. % More implementation details are in the appendix.

In the experiments, the dropout rate is $0.1$. We set $\lambda_{IT},\lambda_{C},\lambda_{U}=1.0, 0.05, 0.5$ in Eq.~(\ref{overall_loss}), $\mu,\gamma=0.6,0.25$ in Eq.~(\ref{pseudo_link_labels}). The batch size is set as $32$. Adam~\cite{kingma2014adam} with $\beta_{1}=0.9$, $\beta_{2}=0.999$ is used as the optimizer. We use the linear-decay learning rate scheduler with 10\% warmup, and train the mode with 15 and 20 epochs ($t_{max}$) on MS-COCO and Flickr30K, respectively. 

During knowledge discovery, for computing image (sentence) popularity, we sample the counterpart sentences (images) which earn top-$K$ tiers. In MS-COCO, we set $K=7, 2$ and the power parameter $\mu=0.98, 1.0$ in Eq.~(\ref{improved_version_pseudo_linking}) for image and sentence, respectively. While in Flickr30K, we set $K=10, 2$ and power parameter $\mu=0.96, 1.0$. For ablation study (\S\ref{section_ablation_study}) and other experiments (\S\ref{section_discussion}), we set the hyperparameters of $K$ and $\mu$ as the same with that in popularity computation for equal comparison. 

For computation efficiency, during the knowledge discovery, instead of computing the linking confidence score for each pair across the unimodal datasets, which needs $\#\mathcal{T}\times\#\mathcal{I}$ times of calculation, we firstly perform cross-modal retrieval with CLIP and select the top 40 candidate sentences (images) for each image (sentence), then the knowledge discovery of each image or sentence is limited in the corresponding 40 pairs. The number of calculation is reduced to $(\#\mathcal{T}+\#\mathcal{I})\times 40$.

\subsection{Performance Comparison}
\textbf{Compared Baselines.}\ \ We compare the proposed framework with several vision language pre-trained models, including CLIP(ViT-B/32)~\cite{radford2021learning} and VinVL(base)~\cite{zhang2021vinvl}. For knowledge discovery, in addition to CLIP and VinVL, we also compare with two weakly supervised phase grounding models, i.e., CLPG~\cite{gupta2020contrastive} and CKD~\cite{wang2021improving} on Flickr30K in terms of local level link prediction. For Image-text retrieval, since our framework does not require any paired image-text data for training, for a fair comparison, the CLIP is evaluated under zero-shot setting, and VinVL is trained with the links obtained from CLIP.

%In image-sentence global level linking, We compare our model with methods based on vision language pre-trained models: CLIP(ViT-B/32)~\cite{radford2021learning} and VinVL(base)~\cite{zhang2021vinvl}.
% UNITER~\cite{chen2020uniter}, CLIP(ViT-B/32)~\cite{radford2021learning}, ALIGN~\cite{jia2021scaling}, and ALBEF~\cite{li2021align}. 
%Without the awareness of the training set, we show the zero-shot performance of CLIP, train VinVL with the pseudo labels produced by CLIP, and utilize it to construct the graph following the absolute threshold based linking in Eq.~(\ref{base_version_pseudo_linking}). 

\textbf{Results and Analysis.} Table~\ref{OverallPerformance} summarizes the performance comparison results. For knowledge discovery, our method achieves remarkable improvement compared with CLIP and VinVL for both global level and local level link prediction, showing the effectiveness of our method in knowledge discovery. It is worthwhile to mention that our method also outperforms weakly-supervised phase grounding methods CLPG and CKD for local level link prediction without any supervision. For image-text retrieval, similar observations can be founded. Our method outperforms both CLIP and VinVL, attaining the best performances in terms of image-text retrieval. The superior performances attained on both knowledge discovery and image-text retrieval demonstrate the effectiveness of the proposed iterative vision-text pre-training framework.

\begin{table*}[ht]
\caption{Overall performance in the test set of Flickr30K~\cite{young2014image} and MS-COCO~\cite{lin2014microsoft}.}
\label{OverallPerformance}
\begin{center}
% \resizebox{\textwidth}{!}{
\begin{tabular}{lcccc|cccccc}
    \toprule
        &\multicolumn{4}{c|}{Knowledge Discovery} &\multicolumn{6}{c}{Image-text Retrieval} \\
        \midrule[0.5pt]
        \multirow{2}{*}{Model} 
        &\multicolumn{3}{c}{Global} &Local &\multicolumn{3}{c}{Image-to-Sentence} &\multicolumn{3}{c}{Sentence-to-Image} \\
        &R &P &F &Acc &R@1 &R@5 &R@10 &R@1 &R@5 &R@10\\
        % \midrule[1.0pt]
        \midrule[1.0pt]
        &\multicolumn{10}{c}{Flickr30K}\\
        % &\multicolumn{7}{c}{Zero-shot}\\
        % \midrule[0.3pt]
        \midrule
        % \emph{UNITER}~\cite{chen2020uniter} &83.6 &95.7 &97.7 &68.7 &89.2 &93.9 \\
        \emph{CLPG}~\cite{gupta2020contrastive} &- &- &-  &51.7 &- &- &- &- &- &- \\
        \emph{CKD}~\cite{wang2021improving} &- &- &- &53.1 &- &- &- &- &- &- \\
        % \emph{ALIGN}~\cite{jia2021scaling} &88.6 &98.7 &99.7 &75.7 &93.8 &96.8 &553.3 &- &- &- &- &- \\
        \emph{CLIP}~\cite{radford2021learning} &42.6 &54.1 &47.7 &-  &79.2 &94.7 &98.3 &58.6 &83.4 &89.9 \\ % &540.6 % 0.32 3941 0.5409794468409034 0.4264
        % 1 0.792 1 0.5858 5 0.947 5 0.834 10 0.983 10 0.8994
        % \emph{ALBEF}~\cite{li2021align} &90.5 &98.8 &99.7 &76.8 &93.7 &96.7 \\
        % \midrule[1.0pt]
        % \midrule
        % &\multicolumn{7}{c}{Transductive Unsupervised Learning}\\
        % \midrule[0.3pt]
        % \midrule
        \emph{VinVL}~\cite{zhang2021vinvl} &64.1 &80.3 &71.3 &50.4 &89.1 &99.1 &99.7 &75.1 &93.5 &96.1  \\ % &555.4 % 7369
        \emph{Ours}  &\textbf{72.2} &\textbf{88.1} &\textbf{79.4} &\textbf{58.7} &\textbf{94.3} &\textbf{99.5} &\textbf{99.7} &\textbf{82.4} &\textbf{95.0} &\textbf{96.5}  \\  % &567.4 
        \midrule[1.0pt]
        &\multicolumn{10}{c}{MS-COCO}\\
        % \cite{parcalabescu2020exploring} &- &- &- &- &- &- &- &- &- &- &- &56.3 \\
        % \cite{wang2019phrase} &- &- &- &- &- &- &- &- &- &- &- &57.1 \\
        % \emph{ALIGN}~\cite{jia2021scaling} &88.6 &98.7 &99.7 &75.7 &93.8 &96.8 &553.3 &- &- &- &- &- \\
        \midrule
        \emph{CLIP}~\cite{radford2021learning} &11.7 &28.7 &16.7 &-  &50.2 &75.0 &83.6 &30.4 &56.0 &66.9 \\ % &540.6 & 10256 0.11764 0.28675897035881437 0.16683685046516905
        % \emph{ALBEF}~\cite{li2021align} &90.5 &98.8 &99.7 &76.8 &93.7 &96.7 \\
        % \midrule[1.0pt]
        % \midrulefgh 
        % &\multicolumn{7}{c}{Transductive Unsupervised Learning}\\
        % \midrule[0.3pt]
        % \midrule
        \emph{VinVL}~\cite{zhang2021vinvl} &29.1 &70.7 &41.2  &-  &65.8 &86.3 &90.8 &50.2 &75.2 &81.7\\ % 7269
        \emph{Ours} &\textbf{33.6} &\textbf{79.5} &\textbf{47.2}  &- &\textbf{71.5} &\textbf{87.2} &\textbf{91.7} &\textbf{54.0} &\textbf{76.5} &\textbf{82.3} \\ %&567.1 % 8392
        % \midrule
        % &\multicolumn{7}{c}{Inductive Supervised Learning}\\
        % \midrule[0.3pt]
        % \emph{VinVL}~\cite{zhang2021vinvl} &94.2 &99.6 &99.7 &82.9 &96.6 &97.9 &570.9 \\
    \bottomrule
\end{tabular}
% }
\end{center}
\end{table*}

%Compare with CLIP and VinVL, our proposed model achieves remarkable improvement. In image-sentence level linking, our model improves the the recall, precision and f1 on the basis of VinVL. The model performance in image-text retrieval also achieves best performance in terms of all metrics. In concept level, the accuracy of our method is 7.7 points higher than VinVL, and this indicates that our framework improves the capability in fine-grained linking and proves that the quality of concept-level links in our graph. 

\subsection{Ablation Study}
\label{section_ablation_study}
We further investigate the effectiveness of different modules in our framework, which includes knowledge discovery (KD), confidence score aware loss (CAL), graph structure enhanced representation learning (GL) and uncertainty reducing (UR) in multi-modal pre-training. Table~\ref{AblationPerformance} summarizes the ablation study results on Flickr30K using VinVL as the base model. From the results, we have the following observations. First, when adding each of the modules, the performances of both knowledge discovery and image-text retrieval have been improved, which suggests the effectiveness of these components. Second, it achieves the best performance when using all components, demonstrating these four modules are complementary to each other.

\begin{table*}[ht]
\caption{Ablation study in the test set of Flickr30K~\cite{young2014image}. } %IT and C is short for link prediction in image-sentence level and concept level.} 
\label{AblationPerformance}
\begin{center}
% \resizebox{1.0\textwidth}{!}{
\begin{tabular}{lcccc|cccccc}
    \toprule
        &\multicolumn{4}{c|}{Knowledge Discovery} &\multicolumn{6}{c}{Image-text Retrieval} \\
        \midrule[0.5pt]
        \multirow{2}{*}{Model} 
        &\multicolumn{3}{c}{Global} &Local &\multicolumn{3}{c}{Image-to-Sentence} &\multicolumn{3}{c}{Sentence-to-Image}\\
        &R &P &F &Acc &R@1 &R@5 &R@10 &R@1 &R@5 &R@10 \\
        % \midrule[1.0pt]
        \midrule[1.0pt]
        % \midrule
        % &\multicolumn{7}{c}{Transductive Unsupervised Learning}\\
        % \midrule[0.3pt]
        % \midrule
        \emph{VinVL}  &64.1 &80.3 &71.3 &50.4 &89.1 &99.1 &99.7 &75.1 &93.5 &96.1 \\ %3207
        \emph{$+$KD} &68.0 &83.9 &75.1 &55.1 &91.1 &99.1 &99.7 &77.8 &94.2 &96.2  \\ %3402
        \emph{$+$CAL} &70.8 &85.7 &77.5 &56.8 &92.8 &99.1 &99.7 &80.1 &94.6 &96.3  \\ %3504
        \emph{$+$GL} &71.3  &86.6 &78.2 &57.6 &93.6 &99.1 &99.7 &81.7 &94.8 &96.5 \\ %3569
        \emph{$+$UR} &\textbf{72.2}  &\textbf{88.1} &\textbf{79.4} &\textbf{58.7} &\textbf{94.3} &\textbf{99.5} &\textbf{99.7} &\textbf{82.4} &\textbf{95.0} &\textbf{96.5} \\ %3611
    \bottomrule
\end{tabular}
% }
\end{center}
\end{table*}

\subsection{Discussion}
\label{section_discussion}
\begin{figure}[!th]
    \centering
    \subfigure[Performance of global level knowledge discovery with respect to iteration $t$. X-axis is the iteration step $t$, Y-axis is the score of precision, recall and f1.]{\includegraphics[width=0.8\textwidth]{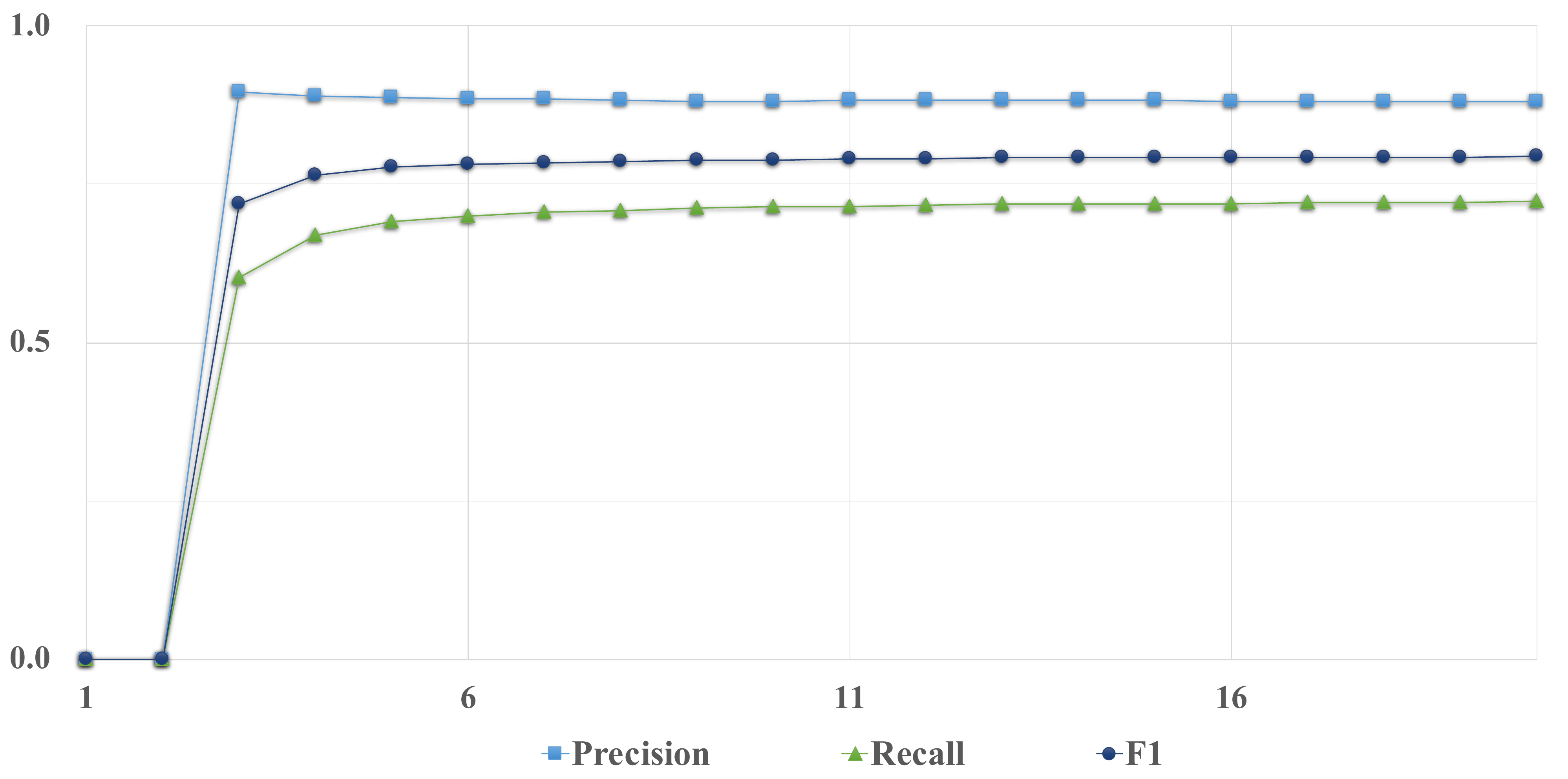}} 
    \subfigure[Performance in image-text retrieval with respect to iteration $t$. X-axis is the iteration step $t$, Y-axis is the score of R@1.]{\includegraphics[width=0.8\textwidth]{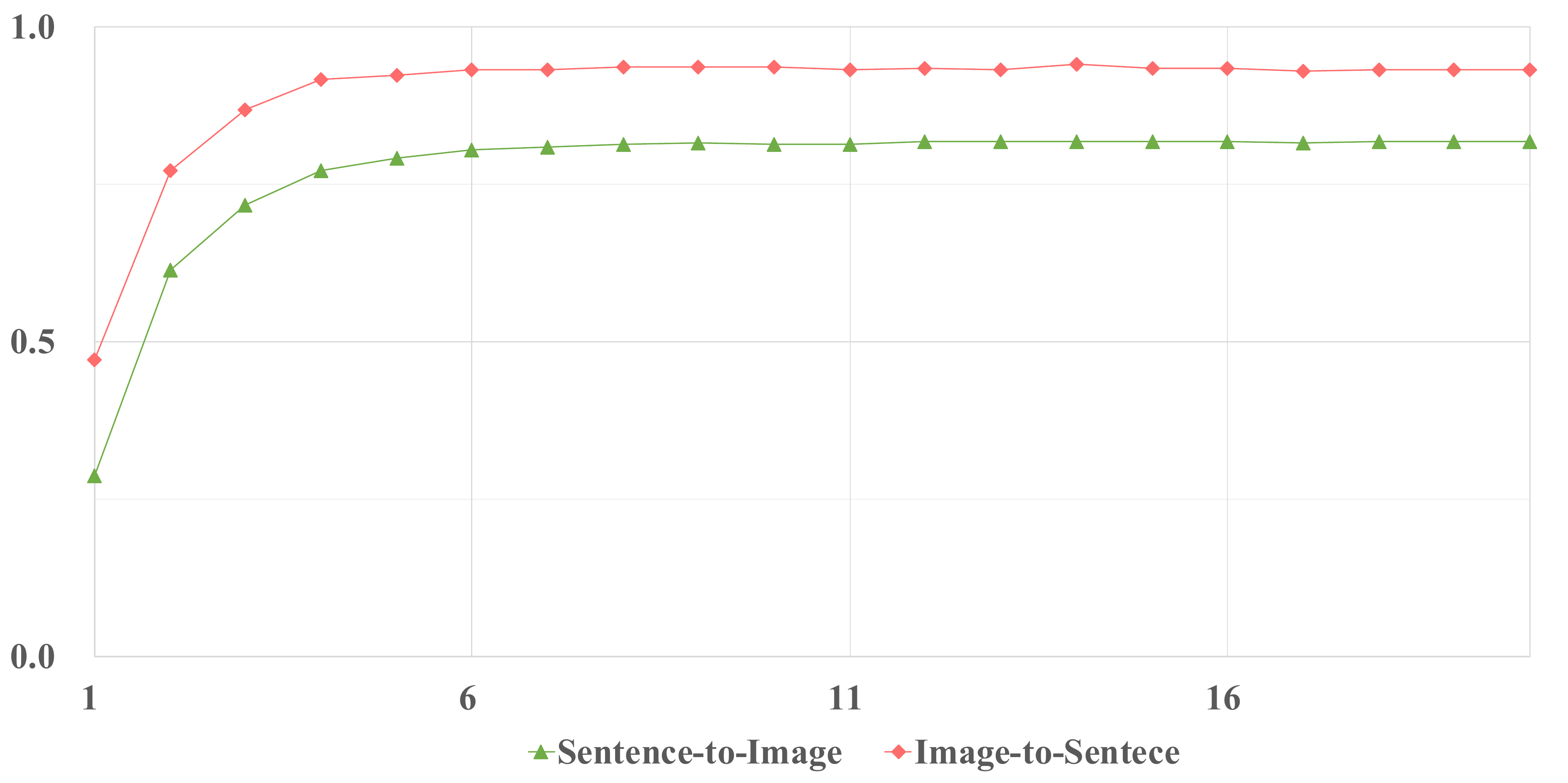}} 
    \caption{Evaluation of each iteration step of the continuous learning framework in the test set of Flickr30K~\cite{plummer2015flickr30k}.}
    \label{iteration_result}
\end{figure}

\paragraph{Continuous Learning Performance w.r.t Iteration Step.} In the test set of Flickr30K~\cite{plummer2015flickr30k}, we evaluate each iteration step of the framework to validate the effectiveness of the proposed continuous learning strategy. The performance with respect to global level knowledge discovery and image-text retrieval is shown in Figure~\ref{iteration_result}. From the table, we can see that as the iteration goes on, the overall performance in both tasks gets better in the early stage and is maintained in the later. This demonstrates that (1) the knowledge discovery and the multi-modal model can benefit each other for successive improvement; (2) despite the error in knowledge discovery is inevitable during the iteration, our framework with confidence score aware training can alleviate this issue hence preventing the failure of training. 

\paragraph{Effects of Different Link Assignment Strategies.} As mentioned in \S\ref{Global_Level_Link_Prediction}, we have introduced different link assignment strategies in the process of knowledge discovery. These strategies include (1) the comparison one, absolute threshold based link assignment (AT)~\cite{lee2013pseudo,shi2018transductive} in Eq.~(\ref{base_version_pseudo_linking}), (2) label-aware link assignment (LA) and bi-label link assignment (BL) that can be successively added to VinVL. To investigate the effects of different strategies, we conduct experiments on Flickr30K datasets and the experimental results are reported in Table~\ref{LinkingStrategyPerformance}. To demonstrate that the proposed link assignment strategies are also effective in dealing with the imbalance problem in link prediction, we report the proportion of the links from popular nodes (\%PP) that have more than 10 links. 
From the results, we find that: (1) The problem of imbalance is serious in VinVL and VinVL+AT. Through integrating LA and BL into VinVl, no nodes will have more than 10 links as the \%PP is reduced to 0. (2) Compared to VinVL and AT, LA and BL successively improve the performance of link prediction as well as the image-text retrieval. % By combining AT with BL, it achieves the best performances on both knowledge discovery and image-text retrieval.  % As mentioned in \S\ref{Global_Level_Link_Prediction}, we have introduced different link assignment strategies in the process of knowledge discovery. These strategies include (1) the classic absolute threshold based link assignment (AT)~\cite{lee2013pseudo,shi2018transductive} in Eq.~(\ref{base_version_pseudo_linking}), (2) label-aware link assignment (LA) and bi-label link assignment (BL). To investigate the effects of different strategies, we conduct experiments on Flickr30K datasets and the experimental results are reported in Table~\ref{LinkingStrategyPerformance}. To demonstrate that the proposed link assignment strategies is also effective in dealing with the imbalance problem in link prediction, we also report the proportion of the links from popular nodes (\%PP) that has more than 10 links. From the results, we find that: (1) the proposed link assignment strategies are all effective in handle the imbalance problem, by combining AT with BL, no nodes will have more than 10 links as the \%PP is reduced to 0. (2) Compared to LA, BL is more effective in improving the performance of link prediction as well as image-text retrieval. By combining AT with BL, it achieves the best performances on both knowledge discovery and image-text retrieval.  

\begin{table*}[ht]
\caption{Performances of different link assignment strategies on Flickr30K~\cite{young2014image}. VinVL is used as the base model. Confidence score aware loss is adopted when using different link assignment strategies.} %IT and C is short for link prediction in image-sentence level and concept level.} 
\label{LinkingStrategyPerformance}
\begin{center}
% \resizebox{1.0\textwidth}{!}{
\begin{tabular}{lcccc|cccccc}
    \toprule
        &\multicolumn{4}{c|}{Knowledge Discovery} &\multicolumn{6}{c}{Image-text Retrieval} \\
        \midrule[0.5pt]
        \multirow{2}{*}{Model} 
        &\multicolumn{4}{c|}{Global} &\multicolumn{3}{c}{Image-to-Sentence} &\multicolumn{3}{c}{Sentence-to-Image}\\
        &\%PP &R &P &F &R@1 &R@5 &R@10 &R@1 &R@5 &R@10 \\
        % \midrule[1.0pt]
        \midrule[1.0pt]
        % \midrule
        % &\multicolumn{7}{c}{Transductive Unsupervised Learning}\\
        %         \emph{$+$PGC} &92.1 &99.1 &99.7 &78.4 &94.2 &96.2 &4,053 &68.0 &83.9 &75.1 &55.1 \\ %3402
        % \midrule[0.3pt]
        % \midrule
        \emph{VinVL} &30.7 &64.1 &80.3 &71.3 &89.1 &99.1 &99.7 &75.1 &93.5 &96.1 \\ %3207
        \midrule[0.3pt]
        \emph{$+$AT} &25.3 &66.3 &81.8 &73.2 &89.6 &98.9 &99.6 &75.9 &93.6 &96.1  \\ %3313
        \midrule[0.3pt] 
        % \emph{$+$PGC} &92.1 &99.1 &99.7 &78.4 &94.2 &96.2 &4,053 &68.0 &83.9 &75.1\\ 
        \emph{$+$LA}&3.8  &\textbf{74.5} &73.9 &74.2 &91.3 &99.1 &99.7 &78.3 &94.0 &96.2  \\ %3724, 5042
        \emph{$+$BL}&\textbf{0.0} &70.8 &\textbf{85.7} &\textbf{77.5} &\textbf{92.8} &\textbf{99.1} &\textbf{99.7} &\textbf{80.1} &\textbf{94.6} &\textbf{96.3}  \\ %3504
    \bottomrule
\end{tabular}
% }
\end{center}
\end{table*}

\paragraph{Performance of Inductive Learning.}
\label{section_inductive_learning}
We then test the generalization ability of our pre-trained model under inductive learning where neither the pseudo graph nor the graph modeling is available. In our implementation, we randomly sample 1,000 images and the corresponding 5,000 sentences from the training set of Flickr30K to construct two unimodal datasets for training, and test the model in the Flickr30K. For comparison, we also train the CLIP~\cite{radford2021learning} and VinVL~\cite{zhang2021vinvl} with the pairs under supervised setting. Supervised VinVL can be regarded as the upper bound. The results are summarized in Table~\ref{InductivePerformance}. 

From the results, our model outperforms VinVL in unsupervised setting and the supervised CLIP, and achieve comparable performance with supervised VinVL. The results basically suggest that our method generalizes well under inductive learning. % There is still a gap between ours and the supervised VinVL, which is a great challenge in the future.
\begin{table*}[ht]
\caption{Performance of multi-modal model in the Flickr30K testing set with inductive manner.} 
\label{InductivePerformance}
\begin{center}
% \resizebox{0.67\textwidth}{!}{
\begin{tabular}{lccccccc}
    \toprule
        \multirow{2}{*}{Model} 
        &\multicolumn{3}{c}{Image-to-Sentence} &\multicolumn{3}{c}{Sentence-to-Image} \\
        &R@1 &R@5 &R@10 &R@1 &R@5 &R@10 \\
        % \midrule[1.0pt]
        % \midrule[1.0pt]
        % &\multicolumn{11}{c}{Flickr30K}\\
        \midrule
        &\multicolumn{6}{c}{Supervised Learning}\\
        \midrule
        % \emph{CLIP}~\cite{radford2021learning} &85.4 &97.0 &98.8 &71.5 &91.6 &95.8 \\ % &540.6 % 0.32 3941 0.5409794468409034 0.4264 0.8519999980926514, 't2i_r1': 0.6869999766349792, 'i2t_r5': 0.9660000205039978, 't2i_r5': 0.8974000215530396, 'i2t_r10': 0.9900000095367432, 't2i_r10': 0.9477999806404114, 'rsum': 5.3402000069618225
        % \emph{CLIP}~\cite{radford2021learning} &85.1 &96.6 &99.0 &68.7 &89.7 &94.8 \\
        \emph{CLIP} &85.1 &96.6 &99.0 &68.7 &89.7 &94.8 \\
        % \emph{VinVL}~\cite{zhang2021vinvl} &93.0 &99.5 &99.9 &79.6 &95.2 &97.1 \\ % &555.4 % 7369
        % \emph{VinVL}~\cite{zhang2021vinvl} &90.3 &99.4 &99.8 &79.4 &94.6 &96.5 \\
        \emph{VinVL} &90.3 &99.4 &99.8 &79.4 &94.6 &96.5 \\
        % 'eval_i2t_r1': 0.9039999842643738, 'eval_t2i_r1': 0.7940000295639038, 'eval_i2t_r5': 0.9940000176429749, 'eval_t2i_r5': 0.946399986743927, 'eval_i2t_r10': 0.9980000257492065, 'eval_t2i_r10': 0.9646000266075134, 'eval_rsum': 5.601000070571899
        \midrule
        &\multicolumn{6}{c}{Unsupervised Learning}\\
        \midrule
        \emph{CLIP} &79.2 &94.7 &98.3 &58.6 &83.4 &89.9 \\ 
        % 1 0.77 1 0.5894 5 0.953 5 0.8478 10 0.984 10 0.9092
        \emph{VinVL} &84.9 &98.4 &99.5 &70.2 &92.0 &95.3 \\ 
        % 'eval_i2t_r1': 0.8489999771118164, ' eval_t2i_r1': 0.7021999955177307, 'eval_i2t_r5': 0.984000027179718, 'eval_t2i_r5': 0.920199990272522, 'eval_i2t_r10': 0.9950000047683716, 'eval_t2i_r10': 0.9531999826431274, 'eval_rsum': 5.403599977493286,
        \emph{Ours} &\textbf{91.2} &\textbf{99.5} &\textbf{99.8} &\textbf{78.3} &\textbf{94.1} &\textbf{96.1} \\ % ''eval_i2t_r1': 0.9129999876022339, 'eval_t2i_r1': 0.7832000255584717, 'eval_i2t_r5': 0.9950000047683716, 'eval_t2i_r5': 0.9413999915122986, 'eval_i2t_r10': 0.996999979019165, 'eval_t2i_r10': 0.9610000252723694
    \bottomrule
\end{tabular}
% }
\end{center}
\end{table*}

\paragraph{Noisy Environment.}
To test the noise-resistant learning capability of the multi-modality model, we setup the noisy environment where not each sample in the uni-modal dataset has a corresponding counterpart. The clean version of unimodal datasets is composed of 500 images which are randomly sampled from the test set of Flickr30K and their corresponding 2,500 sentences. Then, we add noises through (1) randomly adding 2,500 sentences of MS-COCO to the sentence based unimodal dataset; (2) adding 500 images from validation set of Flickr30K to the image based unimodal dataset; (3) both (1) and (2). The noisy samples do not have ground-truth links. We test it on the whole test set of Flickr30K with 1,000 images and 5,000 sentences. We list the results in Table~\ref{synthetic_dataset_result}. 

From the table, we find that: (1) Compare to adding noise to either image or sentence based unimodal dataset, adding noise to both fails to predict another 2\% ground-truth links. (2) Despite the degradation in the quality of the pseudo graph, our noisy-resistant model training is comparable among different settings. This validates the learning efficiency of our framework in the noisy environment.

\begin{table*}[ht]
\caption{Performance comparison in the noisy environment. Noise-$i$ is corresponding to the $i$-th noise adding strategy in the section of Noisy Environment.} % \#I is the number of image sampled from Flickr30K, \#T is the number of sentence sampled from MS-COCO. Sentences from MS-COCO and 2,500 sentences from Flickr30K jointly compose the sentence based unimodal dataset.} 
\label{synthetic_dataset_result}
\begin{center}
\resizebox{0.9\textwidth}{!}{
\begin{tabular}{lccc|cccccc}
    \toprule
        \multirow{2}{*}{Data}
        &\multicolumn{3}{c}{Global} &\multicolumn{3}{c}{Image-to-Sentence} &\multicolumn{3}{c}{Sentence-to-Image}\\
        &R &P &F &R@1 &R@5 &R@10 &R@1 &R@5 &R@10 \\
        % \midrule[1.0pt]
        \midrule[1.0pt]
        % \midrule
        % &\multicolumn{7}{c}{Transductive Unsupervised Learning}\\
        % \midrule[0.3pt]
        % \midrule
        \emph{Clean} &\textbf{76.8} &\textbf{89.9} &\textbf{82.8} &91.5 &99.3 &99.6 &77.8 &\textbf{94.5} &96.2 \\ % 'eval_i2t_r1': 0.9049999713897705, 'eval_t2i_r1': 0.7875999808311462, 'eval_i2t_r5': 0.9929999709129333, 'eval_t2i_r5': 0.9448000192642212, 'eval_i2t_r10': 0.9959999918937683, 'eval_t2i_r10': 0.9617999792098999, 'eval_rsum': 5.5881999135017395, 1921
        \emph{Noise-1} &74.3 &89.5 &81.2 &92.2 &99.4 &\textbf{99.7} &\textbf{78.1} &94.0 &96.1 \\ %1,831, 'eval_i2t_r1': 0.921999990940094, 'eval_t2i_r1': 0.7807999849319458, 'eval_i2t_r5': 0.9940000176429749, 'eval_t2i_r5': 0.9398000240325928, 'eval_i2t_r10': 0.996999979019165, 'eval_t2i_r10': 0.9613999724388123, 'eval_rsum': 5.594999969005585, 2096, 1876
        \emph{Noise-2} &74.3 &86.7 &79.2  &\textbf{92.5} &\textbf{99.4} &99.6 &77.9 &94.3 &\textbf{96.3} \\ 
        % ''eval_i2t_r1': 0.925000011920929, 'eval_t2i_r1': 0.7788000106811523, 'eval_i2t_r5': 0.9940000176429749, 'eval_t2i_r5': 0.9431999921798706, 'eval_i2t_r10': 0.9959999918937683, 'eval_t2i_r10': 0.9629999995231628, 'eval_rsum': 5.600000023841858, 'eval_pseudo_linking_number': 2143, 'eval_pseudo_linking_correct_number': 1858
        \emph{Noise-3} &72.6 &82.1 &77.1  &92.4 &99.3 &99.7 &77.0 &94.1 &96.1 \\ %4027, 'eval_pseudo_linking_correct_number': 3479 'eval_i2t_r1': 0.9240000247955322, 'eval_t2i_r1': 0.7698000073432922, 'eval_i2t_r5': 0.9929999709129333, 'eval_t2i_r5': 0.9409999847412109, 'eval_i2t_r10': 0.996999979019165, 'eval_t2i_r10': 0.9611999988555908, 'eval_rsum': 5.585999965667725, eval_pseudo_linking_number': 2210, 'eval_pseudo_linking_correct_number': 1816, 'eval_pseudo_linking_correct_rate': 0.6957854406130268, 'eval_pseudo_linking_error_number': 794, 'eval_pseudo_linking_error_rate': 0.30421455938697317,
    \bottomrule
\end{tabular}
}
\end{center}
\end{table*}

\paragraph{Random Uni-modal Datasets where No Ground-truth Link Exists.}
To demonstrate that our framework does not heavily rely on the selection of the unimodal datasets, we setup a random environment where none of image or sentence has ground-truth counterpart. In the environment, instead of selecting both of image and text based uni-modal datasets from Flickr30K testing set in \S\ref{section_experiment_setup}, we keep the 1,000 images of Flickr30K testing set as the image based uni-modal dataset, and pick up 5,000 sentences from MS-COCO training set to compose the text based uni-modal dataset. In evaluation, considering that no ground-truth link exists in the random environment, we ignore the task of knowledge discovery and only report the performance of image-text retrieval in Flickr30K testing set. The first three comparison models are trained with the uni-modal datasets where both of the image and text based uni-modal datasets from the Flickr30K testing set. The result is listed in Table~\ref{random_result}. From the table, we find that the missing of ground-truth counterpart degrades the model performance in image-text retrieval. Compare with VinVL(Random) and the counterpart model VinVL, Ours(Random) has less performance drop from its counterpart. The performance of Ours(Random) is better than CLIP and comparable with VinVL which are trained with the uni-modal datasets where the ground-truth link exists. This demonstrates that our framework is effective when training on random unimodal datasets.
% the image based uni-modal dataset is the 1,000 images of Flickr30K testing set,
\begin{table*}[ht]
\caption{Performance comparison in the random environment where none of image or sentence has ground-truth counterpart. The first three comparison models are trained with the uni-modal datasets where the ground-truth link exists.}
\label{random_result}
\begin{center}
%\resizebox{0.9\textwidth}{!}{
\begin{tabular}{lcccccc}
    \toprule
        \multirow{2}{*}{Model}
        &\multicolumn{3}{c}{Image-to-Sentence} &\multicolumn{3}{c}{Sentence-to-Image}\\
        &R@1 &R@5 &R@10 &R@1 &R@5 &R@10 \\
        % \midrule[1.0pt]
        \midrule[1.0pt]
        % \midrule
        % &\multicolumn{7}{c}{Transductive Unsupervised Learning}\\
        % \midrule[0.3pt]
        % \midrule
        \emph{CLIP}~\cite{radford2021learning} &79.2 &94.7 &98.3 &58.6 &83.4 &89.9 \\ % 
        \emph{VinVL} &89.1 &99.1 &99.7 &75.1 &93.5 &96.1 \\
        \emph{Ours}  &92.2 &99.4 &99.7 &78.1 &94.0 &96.1 \\ %1,831, 
        \midrule
        \emph{VinVL(Random)} &75.9 &98.1 &99.5 &66.8 &92.2 &95.6 \\ % 'eval_i2t_r1': 0.7590000033378601, 'eval_t2i_r1': 0.66839998960495, 'eval_i2t_r5': 0.9810000061988831, 'eval_t2i_r5': 0.9218000173568726, 'eval_i2t_r10': 0.9950000047683716, 'eval_t2i_r10': 0.9557999968528748, 'eval_rsum': 5.281000018119812
        \emph{Ours(Random)}  &89.7 &99.3 &99.7 &75.7 &93.2 &96.0 \\ % 'eval_i2t_r1': 0.8970000147819519, 'eval_t2i_r1': 0.7174000144004822, 'eval_i2t_r5': 0.9929999709129333, 'eval_t2i_r5': 0.921999990940094, 'eval_i2t_r10': 0.996999979019165, 'eval_t2i_r10': 0.9556000232696533, 'eval_rsum': 5.48199999332428
    \bottomrule
\end{tabular}
% }
\end{center}
\end{table*}

\section{Conclusion}
In this paper, we propose a unified continuous learning framework for multi-modal knowledge discovery and pre-training based on two uni-modal datasets. In the framework, knowledge discovery is regarded as the problem of inter-modality link prediction on a multi-modal knowledge graph. Multi-modal pre-training is then performed with the guidance of the knowledge graph. In order to mitigate the error propagation problem in the iterative process, we propose label-aware link assignment strategy based on the popularity of nodes in the graph construction. For model pre-training, we take graph structure as the knowledge for guidance. In practice, two-hop neighbors are utilized for the node representation learning. The experiments on Flickr30K and MS-COCO validate the quality of the multi-modal graph and the performance of pre-training model. The further analysis in the noisy environment also demonstrates that our system can perform reliable model training. 

There are three major future directions. Firstly, more pre-training tasks on the multi-modal graph can be introduced to our framework for better model pre-training. Secondly, it would be interesting to explore a more efficient way for better continuous learning with efforts on quality control of the multi-modal knowledge graph. Thirdly, we would like to explore the structure of multi-modal graph for better cross-modality knowledge representation. 

% In the process of link prediction, we utilize the pre-trained one-tower model VinVL to output the confidence score. Despite that we reduce the number of the possible linking counterpart to 40 with CLIP, and the computation is heavy. With 4 24GB 3090, building the pseudo multi-modality graph through link prediction over the 5k images and 25k sentences unimodal datasets requires about 30 minutes. Thus, the time-consuming is heavy. 

% configure the concept-enhanced cross-modal graph to explore the global association between concepts and other sentences. In the configuration, we set the concepts as cornerstones to connect with both images and sentences. And to apply the graph in image-text retrieval, we regard it as link prediction. In the following graph encoding, we sample sub-graph and define the shortest cross-modal meta-path to capture the topological feature. Meta-path based transformer is proposed for graph structure modeling. We also perform concept link prediction as auxiliary task. Experiment results show the effectiveness of our model and further analysis reveals that our meta-path based transformer has powerful capability in graph structure modeling. In the future, we would like to combine the configuration of the concept-enhanced cross-modal graph and the previous denotation graph which follows the linguistic expression to build the hierarchical relationship within concepts. 

% \section*{References}
\bibliographystyle{unsrtnat}
\bibliography{neurips_2022,copy}

\begin{thebibliography}{49}
\providecommand{\natexlab}[1]{#1}
\providecommand{\url}[1]{\texttt{#1}}
\expandafter\ifx\csname urlstyle\endcsname\relax
  \providecommand{\doi}[1]{doi: #1}\else
  \providecommand{\doi}{doi: \begingroup \urlstyle{rm}\Url}\fi

\bibitem[Lin et~al.(2014)Lin, Maire, Belongie, Hays, Perona, Ramanan,
  Doll{\'a}r, and Zitnick]{lin2014microsoft}
Tsung-Yi Lin, Michael Maire, Serge Belongie, James Hays, Pietro Perona, Deva
  Ramanan, Piotr Doll{\'a}r, and C~Lawrence Zitnick.
\newblock Microsoft coco: Common objects in context.
\newblock In \emph{European conference on computer vision}, pages 740--755.
  Springer, 2014.

\bibitem[Young et~al.(2014)Young, Lai, et~al.]{young2014image}
Peter Young, Alice Lai, et~al.
\newblock From image descriptions to visual denotations: New similarity metrics
  for semantic inference over event descriptions.
\newblock \emph{T-ACL}, 2014.

\bibitem[Fan et~al.(2021{\natexlab{a}})Fan, Wei, Li, Wang, and
  Fan]{fan2021negative}
Zhihao Fan, Zhongyu Wei, Zejun Li, Siyuan Wang, and Jianqing Fan.
\newblock Negative sample is negative in its own way: Tailoring negative
  sentences for image-text retrieval.
\newblock \emph{arXiv preprint arXiv:2111.03349}, 2021{\natexlab{a}}.

\bibitem[Fan et~al.(2021{\natexlab{b}})Fan, Wei, Li, Wang, Shan, Huang, and
  Fan]{fan2021constructing}
Zhihao Fan, Zhongyu Wei, Zejun Li, Siyuan Wang, Haijun Shan, Xuanjing Huang,
  and Jianqing Fan.
\newblock Constructing phrase-level semantic labels to form multi-grained
  supervision for image-text retrieval.
\newblock \emph{arXiv:2109.05523}, 2021{\natexlab{b}}.

\bibitem[Antol et~al.(2015)Antol, Agrawal, Lu, Mitchell, Batra, Zitnick, and
  Parikh]{antol2015vqa}
Stanislaw Antol, Aishwarya Agrawal, Jiasen Lu, Margaret Mitchell, Dhruv Batra,
  C~Lawrence Zitnick, and Devi Parikh.
\newblock Vqa: Visual question answering.
\newblock In \emph{Proceedings of the IEEE international conference on computer
  vision}, pages 2425--2433, 2015.

\bibitem[Goyal et~al.(2017)Goyal, Khot, Summers-Stay, Batra, and
  Parikh]{goyal2017making}
Yash Goyal, Tejas Khot, Douglas Summers-Stay, Dhruv Batra, and Devi Parikh.
\newblock Making the v in vqa matter: Elevating the role of image understanding
  in visual question answering.
\newblock In \emph{Proceedings of the IEEE conference on computer vision and
  pattern recognition}, pages 6904--6913, 2017.

\bibitem[Yu et~al.(2016)Yu, Poirson, Yang, Berg, and Berg]{yu2016modeling}
Licheng Yu, Patrick Poirson, Shan Yang, Alexander~C Berg, and Tamara~L Berg.
\newblock Modeling context in referring expressions.
\newblock In \emph{European Conference on Computer Vision}, pages 69--85.
  Springer, 2016.

\bibitem[Plummer et~al.(2015)Plummer, Wang, Cervantes, Caicedo, Hockenmaier,
  and Lazebnik]{plummer2015flickr30k}
Bryan~A Plummer, Liwei Wang, Chris~M Cervantes, Juan~C Caicedo, Julia
  Hockenmaier, and Svetlana Lazebnik.
\newblock Flickr30k entities: Collecting region-to-phrase correspondences for
  richer image-to-sentence models.
\newblock In \emph{Proceedings of the IEEE international conference on computer
  vision}, pages 2641--2649, 2015.

\bibitem[Lu et~al.(2019)Lu, Batra, Parikh, and Lee]{lu2019vilbert}
Jiasen Lu, Dhruv Batra, Devi Parikh, and Stefan Lee.
\newblock Vilbert: Pretraining task-agnostic visiolinguistic representations
  for vision-and-language tasks.
\newblock \emph{Advances in neural information processing systems}, 32, 2019.

\bibitem[Li et~al.(2019)Li, Yatskar, Yin, Hsieh, and Chang]{li2019visualbert}
Liunian~Harold Li, Mark Yatskar, Da~Yin, Cho-Jui Hsieh, and Kai-Wei Chang.
\newblock Visualbert: A simple and performant baseline for vision and language.
\newblock \emph{arXiv preprint arXiv:1908.03557}, 2019.

\bibitem[Su et~al.(2019)Su, Zhu, Cao, Li, Lu, Wei, and Dai]{su2019vl}
Weijie Su, Xizhou Zhu, Yue Cao, Bin Li, Lewei Lu, Furu Wei, and Jifeng Dai.
\newblock Vl-bert: Pre-training of generic visual-linguistic representations.
\newblock \emph{arXiv preprint arXiv:1908.08530}, 2019.

\bibitem[Li et~al.(2020{\natexlab{a}})Li, Yin, Li, Hu, et~al.]{li2020oscar}
Xiujun Li, Xi~Yin, Chunyuan Li, Xiaowei Hu, et~al.
\newblock Oscar: Object-semantics aligned pre-training for vision-language
  tasks.
\newblock \emph{ECCV}, 2020{\natexlab{a}}.

\bibitem[Tan and Bansal(2019)]{tan2019lxmert}
Hao Tan and Mohit Bansal.
\newblock Lxmert: Learning cross-modality encoder representations from
  transformers.
\newblock \emph{arXiv:1908.07490}, 2019.

\bibitem[Li et~al.(2020{\natexlab{b}})Li, Duan, Fang, et~al.]{li2020unicoder}
Gen Li, Nan Duan, Yuejian Fang, et~al.
\newblock Unicoder-vl: A universal encoder for vision and language by
  cross-modal pre-training.
\newblock In \emph{AAAI}, 2020{\natexlab{b}}.

\bibitem[Zhou et~al.(2020)Zhou, Palangi, Zhang, Hu, Corso, and
  Gao]{zhou2020unified}
Luowei Zhou, Hamid Palangi, Lei Zhang, Houdong Hu, Jason Corso, and Jianfeng
  Gao.
\newblock Unified vision-language pre-training for image captioning and vqa.
\newblock In \emph{Proceedings of the AAAI Conference on Artificial
  Intelligence}, volume~34, pages 13041--13049, 2020.

\bibitem[Li et~al.(2021)Li, Selvaraju, Gotmare, Joty, Xiong, and
  Hoi]{li2021align}
Junnan Li, Ramprasaath Selvaraju, Akhilesh Gotmare, Shafiq Joty, Caiming Xiong,
  and Steven Chu~Hong Hoi.
\newblock Align before fuse: Vision and language representation learning with
  momentum distillation.
\newblock \emph{Advances in Neural Information Processing Systems}, 34, 2021.

\bibitem[Kim et~al.(2021)Kim, Son, and Kim]{kim2021vilt}
Wonjae Kim, Bokyung Son, and Ildoo Kim.
\newblock Vilt: Vision-and-language transformer without convolution or region
  supervision.
\newblock In \emph{International Conference on Machine Learning}, pages
  5583--5594. PMLR, 2021.

\bibitem[Zhang et~al.(2021)Zhang, Li, Hu, Yang, Zhang, Wang, Choi, and
  Gao]{zhang2021vinvl}
Pengchuan Zhang, Xiujun Li, Xiaowei Hu, Jianwei Yang, Lei Zhang, Lijuan Wang,
  Yejin Choi, and Jianfeng Gao.
\newblock Vinvl: Revisiting visual representations in vision-language models.
\newblock In \emph{Proceedings of the IEEE/CVF Conference on Computer Vision
  and Pattern Recognition}, pages 5579--5588, 2021.

\bibitem[Li et~al.(2022)Li, Fan, Tou, and Wei]{li2022mvp}
Zejun Li, Zhihao Fan, Huaixiao Tou, and Zhongyu Wei.
\newblock Mvp: Multi-stage vision-language pre-training via multi-level
  semantic alignment.
\newblock \emph{arXiv preprint arXiv:2201.12596}, 2022.

\bibitem[Alberts et~al.(2020)Alberts, Huang, Deshpande, Liu, Cho, Vania, and
  Calixto]{alberts2020visualsem}
Houda Alberts, Teresa Huang, Yash Deshpande, Yibo Liu, Kyunghyun Cho, Clara
  Vania, and Iacer Calixto.
\newblock Visualsem: a high-quality knowledge graph for vision and language.
\newblock \emph{arXiv preprint arXiv:2008.09150}, 2020.

\bibitem[Radford et~al.(2021)Radford, Kim, Hallacy, Ramesh, Goh, Agarwal,
  Sastry, Askell, Mishkin, Clark, et~al.]{radford2021learning}
Alec Radford, Jong~Wook Kim, Chris Hallacy, Aditya Ramesh, Gabriel Goh,
  Sandhini Agarwal, Girish Sastry, Amanda Askell, Pamela Mishkin, Jack Clark,
  et~al.
\newblock Learning transferable visual models from natural language
  supervision.
\newblock In \emph{International Conference on Machine Learning}, pages
  8748--8763. PMLR, 2021.

\bibitem[Chen et~al.(2022)Chen, Zhang, Li, Deng, Tan, Xu, Huang, Si, and
  Chen]{chen2022hybrid}
Xiang Chen, Ningyu Zhang, Lei Li, Shumin Deng, Chuanqi Tan, Changliang Xu, Fei
  Huang, Luo Si, and Huajun Chen.
\newblock Hybrid transformer with multi-level fusion for multimodal knowledge
  graph completion.
\newblock \emph{arXiv preprint arXiv:2205.02357}, 2022.

\bibitem[Vaswani et~al.(2017)Vaswani, Shazeer, Parmar, Uszkoreit, Jones, Gomez,
  Kaiser, and Polosukhin]{vaswani2017attention}
Ashish Vaswani, Noam Shazeer, Niki Parmar, Jakob Uszkoreit, Llion Jones,
  Aidan~N Gomez, \L~ukasz Kaiser, and Illia Polosukhin.
\newblock Attention is all you need.
\newblock In \emph{NeurIPS}. 2017.

\bibitem[Yu et~al.(2020)Yu, Tang, et~al.]{yu2020ernie}
Fei Yu, Jiji Tang, et~al.
\newblock Ernie-vil: Knowledge enhanced vision-language representations through
  scene graph.
\newblock \emph{arXiv:2006.16934}, 2020.

\bibitem[Cui et~al.(2021)Cui, Yu, Wang, Zhao, Zhang, Wang, and
  Yu]{cui2021rosita}
Yuhao Cui, Zhou Yu, Chunqi Wang, Zhongzhou Zhao, Ji~Zhang, Meng Wang, and Jun
  Yu.
\newblock Rosita: Enhancing vision-and-language semantic alignments via
  cross-and intra-modal knowledge integration.
\newblock In \emph{Proceedings of the 29th ACM International Conference on
  Multimedia}, pages 797--806, 2021.

\bibitem[Devlin et~al.(2019)Devlin, Chang, et~al.]{devlin2018bert}
Jacob Devlin, Ming-Wei Chang, et~al.
\newblock Bert: Pre-training of deep bidirectional transformers for language
  understanding.
\newblock In \emph{NAACL}, 2019.

\bibitem[Chen et~al.(2020)Chen, Li, Yu, El~Kholy, Ahmed, Gan, Cheng, and
  Liu]{chen2020uniter}
Yen-Chun Chen, Linjie Li, Licheng Yu, Ahmed El~Kholy, Faisal Ahmed, Zhe Gan,
  Yu~Cheng, and Jingjing Liu.
\newblock Uniter: Universal image-text representation learning.
\newblock In \emph{European conference on computer vision}, pages 104--120.
  Springer, 2020.

\bibitem[Sun et~al.(2019)Sun, Wang, Li, Feng, Chen, Zhang, Tian, Zhu, Tian, and
  Wu]{sun2019ernie}
Yu~Sun, Shuohuan Wang, Yukun Li, Shikun Feng, Xuyi Chen, Han Zhang, Xin Tian,
  Danxiang Zhu, Hao Tian, and Hua Wu.
\newblock Ernie: Enhanced representation through knowledge integration.
\newblock \emph{arXiv preprint arXiv:1904.09223}, 2019.

\bibitem[Zhang et~al.(2019)Zhang, Han, Liu, Jiang, Sun, and
  Liu]{zhang2019ernie}
Zhengyan Zhang, Xu~Han, Zhiyuan Liu, Xin Jiang, Maosong Sun, and Qun Liu.
\newblock Ernie: Enhanced language representation with informative entities.
\newblock \emph{arXiv preprint arXiv:1905.07129}, 2019.

\bibitem[Liu et~al.(2020)Liu, Zhou, Zhao, Wang, Ju, Deng, and Wang]{liu2020k}
Weijie Liu, Peng Zhou, Zhe Zhao, Zhiruo Wang, Qi~Ju, Haotang Deng, and Ping
  Wang.
\newblock K-bert: Enabling language representation with knowledge graph.
\newblock In \emph{Proceedings of the AAAI Conference on Artificial
  Intelligence}, volume~34, pages 2901--2908, 2020.

\bibitem[Wang et~al.(2020)Wang, Tang, Duan, Wei, Huang, Cao, Jiang, Zhou,
  et~al.]{wang2020k}
Ruize Wang, Duyu Tang, Nan Duan, Zhongyu Wei, Xuanjing Huang, Guihong Cao,
  Daxin Jiang, Ming Zhou, et~al.
\newblock K-adapter: Infusing knowledge into pre-trained models with adapters.
\newblock \emph{arXiv preprint arXiv:2002.01808}, 2020.

\bibitem[Wang et~al.(2021{\natexlab{a}})Wang, Gao, Zhu, Zhang, Liu, Li, and
  Tang]{wang2021kepler}
Xiaozhi Wang, Tianyu Gao, Zhaocheng Zhu, Zhengyan Zhang, Zhiyuan Liu, Juanzi
  Li, and Jian Tang.
\newblock Kepler: A unified model for knowledge embedding and pre-trained
  language representation.
\newblock \emph{Transactions of the Association for Computational Linguistics},
  9:\penalty0 176--194, 2021{\natexlab{a}}.

\bibitem[O{\~n}oro-Rubio et~al.(2017)O{\~n}oro-Rubio, Niepert,
  Garc{\'\i}a-Dur{\'a}n, Gonz{\'a}lez, and
  L{\'o}pez-Sastre]{onoro2017answering}
Daniel O{\~n}oro-Rubio, Mathias Niepert, Alberto Garc{\'\i}a-Dur{\'a}n, Roberto
  Gonz{\'a}lez, and Roberto~J L{\'o}pez-Sastre.
\newblock Answering visual-relational queries in web-extracted knowledge
  graphs.
\newblock \emph{arXiv preprint arXiv:1709.02314}, 2017.

\bibitem[Liu et~al.(2019)Liu, Li, Garcia-Duran, Niepert, Onoro-Rubio, and
  Rosenblum]{liu2019mmkg}
Ye~Liu, Hui Li, Alberto Garcia-Duran, Mathias Niepert, Daniel Onoro-Rubio, and
  David~S Rosenblum.
\newblock Mmkg: multi-modal knowledge graphs.
\newblock In \emph{European Semantic Web Conference}, pages 459--474. Springer,
  2019.

\bibitem[Pezeshkpour et~al.(2018)Pezeshkpour, Chen, and
  Singh]{pezeshkpour2018embedding}
Pouya Pezeshkpour, Liyan Chen, and Sameer Singh.
\newblock Embedding multimodal relational data for knowledge base completion.
\newblock \emph{arXiv preprint arXiv:1809.01341}, 2018.

\bibitem[Shah et~al.(2019)Shah, Villmow, Ulges, Schwanecke, and
  Shafait]{shah2019open}
Haseeb Shah, Johannes Villmow, Adrian Ulges, Ulrich Schwanecke, and Faisal
  Shafait.
\newblock An open-world extension to knowledge graph completion models.
\newblock In \emph{Proceedings of the AAAI Conference on Artificial
  Intelligence}, volume~33, pages 3044--3051, 2019.

\bibitem[Hamann et~al.(2021)Hamann, Ulges, Krechel, and
  Bergmann]{hamann2021open}
Felix Hamann, Adrian Ulges, Dirk Krechel, and Ralph Bergmann.
\newblock Open-world knowledge graph completion benchmarks for knowledge
  discovery.
\newblock In \emph{International Conference on Industrial, Engineering and
  Other Applications of Applied Intelligent Systems}, pages 252--264. Springer,
  2021.

\bibitem[Lee et~al.(2013)]{lee2013pseudo}
Dong-Hyun Lee et~al.
\newblock Pseudo-label: The simple and efficient semi-supervised learning
  method for deep neural networks.
\newblock In \emph{Workshop on challenges in representation learning, ICML},
  volume~3, page 896, 2013.

\bibitem[Shi et~al.(2018)Shi, Gong, Ding, Tao, and Zheng]{shi2018transductive}
Weiwei Shi, Yihong Gong, Chris Ding, Zhiheng~MaXiaoyu Tao, and Nanning Zheng.
\newblock Transductive semi-supervised deep learning using min-max features.
\newblock In \emph{Proceedings of the European Conference on Computer Vision
  (ECCV)}, pages 299--315, 2018.

\bibitem[Iscen et~al.(2019)Iscen, Tolias, Avrithis, and Chum]{iscen2019label}
Ahmet Iscen, Giorgos Tolias, Yannis Avrithis, and Ondrej Chum.
\newblock Label propagation for deep semi-supervised learning.
\newblock In \emph{Proceedings of the IEEE/CVF Conference on Computer Vision
  and Pattern Recognition}, pages 5070--5079, 2019.

\bibitem[Rizve et~al.(2021)Rizve, Duarte, Rawat, and Shah]{rizve2021defense}
Mamshad~Nayeem Rizve, Kevin Duarte, Yogesh~S Rawat, and Mubarak Shah.
\newblock In defense of pseudo-labeling: An uncertainty-aware pseudo-label
  selection framework for semi-supervised learning.
\newblock \emph{arXiv preprint arXiv:2101.06329}, 2021.

\bibitem[Girshick(2015)]{girshick2015fast}
Ross Girshick.
\newblock Fast r-cnn.
\newblock In \emph{Proceedings of the IEEE international conference on computer
  vision}, pages 1440--1448, 2015.

\bibitem[Dou and Peng(2021)]{dou2021improving}
Zi-Yi Dou and Nanyun Peng.
\newblock Improving pre-trained vision-and-language embeddings for phrase
  grounding.
\newblock In \emph{Proceedings of the 2021 Conference on Empirical Methods in
  Natural Language Processing}, pages 6362--6371, 2021.

\bibitem[Gal and Ghahramani(2016)]{gal2016dropout}
Yarin Gal and Zoubin Ghahramani.
\newblock Dropout as a bayesian approximation: Representing model uncertainty
  in deep learning.
\newblock In \emph{international conference on machine learning}, pages
  1050--1059. PMLR, 2016.

\bibitem[Yang et~al.(2015)Yang, Lichtenwalter, and Chawla]{yang2015evaluating}
Yang Yang, Ryan~N Lichtenwalter, and Nitesh~V Chawla.
\newblock Evaluating link prediction methods.
\newblock \emph{Knowledge and Information Systems}, 45\penalty0 (3):\penalty0
  751--782, 2015.

\bibitem[Wang et~al.(2021{\natexlab{b}})Wang, Huang, Li, Xu, Yang, and
  Yu]{wang2021improving}
Liwei Wang, Jing Huang, Yin Li, Kun Xu, Zhengyuan Yang, and Dong Yu.
\newblock Improving weakly supervised visual grounding by contrastive knowledge
  distillation.
\newblock In \emph{Proceedings of the IEEE/CVF Conference on Computer Vision
  and Pattern Recognition}, pages 14090--14100, 2021{\natexlab{b}}.

\bibitem[Frome et~al.(2013)Frome, Corrado, Shlens, Bengio,
  et~al.]{NIPS2013_7cce53cf}
Andrea Frome, Greg~S Corrado, Jon Shlens, Samy Bengio, et~al.
\newblock Devise: A deep visual-semantic embedding model.
\newblock In \emph{NeurIPS}, 2013.

\bibitem[Kingma and Ba(2014)]{kingma2014adam}
Diederik~P Kingma and Jimmy Ba.
\newblock Adam: A method for stochastic optimization.
\newblock \emph{arXiv preprint arXiv:1412.6980}, 2014.

\bibitem[Gupta et~al.(2020)Gupta, Vahdat, Chechik, Yang, Kautz, and
  Hoiem]{gupta2020contrastive}
Tanmay Gupta, Arash Vahdat, Gal Chechik, Xiaodong Yang, Jan Kautz, and Derek
  Hoiem.
\newblock Contrastive learning for weakly supervised phrase grounding.
\newblock In \emph{European Conference on Computer Vision}, pages 752--768.
  Springer, 2020.

\end{thebibliography}

\end{document}